\RequirePackage{rotating}
\documentclass[nonblindrev]{informs3_pom}
\OneAndAHalfSpacedXI

\usepackage[draft]{hyperref}% -> Set to draft mode before submission
\RequirePackage{fix-cm} 
\usepackage{xspace}
	\newcommand\ie{i.\,e.\xspace}
	\newcommand\eg{e.\,g.\xspace}

  \newcommand{\norm}[1]{\left\lVert#1\right\rVert}
  
	\usepackage{siunitx}
    \def\sym#1{\ifmmode^{#1}\else\(^{#1}\)\fi}
    \DeclareSIUnit\eur{\officialeuro}
    \DeclareSIUnit\M{M}
    \DeclareSIUnit\k{k}
	
	\usepackage[%
  sort&compress
]{cleveref}
  \crefname{chapter}{section}{sections}
	\Crefname{chapter}{Section}{Sections}

\usepackage{etoolbox}
\robustify\bfseries

\usepackage{isomath}
\usepackage{bm}
\usepackage{enumitem}
\usepackage{colortbl}
\usepackage{tabularx}
\usepackage{longtable}
\usepackage{caption}
\usepackage{booktabs}
\usepackage[flushleft]{threeparttable}
\usepackage{multirow}
\usepackage{lscape}
\usepackage{collcell}
\usepackage{makecell}

\usepackage{etoolbox}           % <--
\renewrobustcmd{\bfseries}{\fontseries{b}\selectfont}
\renewrobustcmd{\boldmath}{}
\newrobustcmd{\B}{\bfseries}

\usepackage{float}
\usepackage{rotating}

\usepackage{array}
\newcolumntype{L}[1]{>{\raggedright\let\newline\\\arraybackslash\hspace{0pt}}p{#1}}
\newcolumntype{C}[1]{>{\centering\let\newline\\\arraybackslash\hspace{0pt}}p{#1}}
\newcolumntype{R}[1]{>{\raggedleft\let\newline\\\arraybackslash\hspace{0pt}}p{#1}}
\usepackage{subcaption}
\usepackage{csquotes}
\usepackage{amsmath}

\usepackage{amssymb}% for \mathbb
\usepackage{pifont}% http://ctan.org/pkg/pifont

\newcommand{\mcellt}[2][c]{%
  \begin{tabular}[t]{@{}#1@{}}#2\end{tabular}}

\usepackage[
    colorinlistoftodos,
    textsize=footnotesize,
        ]{todonotes}

\definecolor{darkgreen}{rgb}{0.0, 0.5, 0.0}

\makeatletter
    \renewcommand{\fps@figure}{H}         % default {tbp}
    \renewcommand{\fps@table}{H}         % default {tbp}
\makeatother 

\usepackage{float}
\usepackage{rotating}
\usepackage{eurosym}

\makeatletter
\newcolumntype{B}[3]{>{\boldmath\DC@{#1}{#2}{#3}}c<{\DC@end}}
\newcolumntype{H}{>{\setbox0=\hbox\bgroup}c<{\egroup}@{}}
\makeatother

\usepackage{natbib}
 \bibpunct[, ]{(}{)}{,}{a}{}{,}%
 \def\bibfont{\small}%
\TheoremsNumberedThrough     % Preferred (Theorem 1, Lemma 1, Theorem 2)
\EquationsNumberedThrough    % Default: (1), (2), ...
\usepackage{eqparbox}

\begin{document}

%%%%%%%%%%%%%%%%
%\renewcommand\thefigure{\arabic{figure}.}
% Outcomment only when entries are known. Otherwise leave as is and 
%   default values will be used.
%\setcounter{page}{1}
%\VOLUME{00}%
%\NO{0}%
%\MONTH{Xxxxx}% (month or a similar seasonal id)
%\YEAR{0000}% e.g., 2005
%\FIRSTPAGE{000}%
%\LASTPAGE{000}%
%\SHORTYEAR{00}% shortened year (two-digit)
%\ISSUE{0000} %
%\LONGFIRSTPAGE{0001} %
%\DOI{10.1287/xxxx.0000.0000}%

% Author's names for the running heads
% Sample depending on the number of authors;
\RUNAUTHOR{Maarouf, Feuerriegel, and Pr\"ollochs}
% \RUNAUTHOR{Jones and Wilson}
% \RUNAUTHOR{Jones, Miller, and Wilson}
% \RUNAUTHOR{Jones et al.} % for four or more authors
% Enter authors following the given pattern:
%\RUNAUTHOR{}

% Title or shortened title suitable for running heads. Sample:
% \RUNTITLE{Bundling Information Goods of Decreasing Value}
% Enter the (shortened) title:
\RUNTITLE{A Fused Large Language Model for Predicting Startup Success}

% Full title. Sample:
% \TITLE{Bundling Information Goods of Decreasing Value}
% Enter the full title:
%\TITLE{Prediction of Successful Startups from VC platforms: The Role of Textual Data in Automated Screening Processes}
%\TITLE{Machine Learning for Predicting Startup Success: \\ The Role of Textual Self-Descriptions on Venture Capital Platforms}
\TITLE{A Fused Large Language Model for Predicting Startup Success}

% Block of authors and their affiliations starts here:
% NOTE: Authors with same affiliation, if the order of authors allows, 
%   should be entered in ONE field, separated by a comma. 
%   \EMAIL field can be repeated if more than one author
\ARTICLEAUTHORS{%
%\AUTHOR{Matthias Gey}
%\AFF{ETH Zurich, \EMAIL{geym@ethz.ch},}
\AUTHOR{Abdurahman Maarouf}
\AFF{LMU Munich \& Munich Center for Machine Learning, \EMAIL{a.maarouf@lmu.de}, \URL{}}
\AUTHOR{Stefan Feuerriegel}
\AFF{LMU Munich \& Munich Center for Machine Learning, \EMAIL{feuerriegel@lmu.de}, \URL{}}
\AUTHOR{Nicolas Pr\"ollochs}
\AFF{Justus Liebig University Giessen, \EMAIL{nicolas.proellochs@wi.jlug.de} \URL{}}

% Enter all authors
} % end of the block

\ABSTRACT{ 
Investors are continuously seeking profitable investment opportunities in startups and, hence, for effective decision-making, need to predict a startup's probability of success. Nowadays, investors can use not only various fundamental information about a startup (e.g., the age of the startup, the number of founders, and the business sector) but also textual description of a startup's innovation and business model, which is widely available through online venture capital (VC) platforms such as Crunchbase. To support the decision-making of investors, we develop a machine learning approach with the aim of locating successful startups on VC platforms. Specifically, we develop, train, and evaluate a tailored, fused large language model to predict startup success. Thereby, we assess to what extent self-descriptions on VC platforms are predictive of startup success. Using 20,172 online profiles from Crunchbase, we find that our fused large language model can predict startup success, with textual self-descriptions being responsible for a significant part of the predictive power. Our work provides a decision support tool for investors to find profitable investment opportunities. 
}

% Fill in data. If unknown, outcomment the field

\maketitle
\sloppy
\raggedbottom
\interfootnotelinepenalty=10000

%%%%%%%%%%%%%%%%%%%%%%%%%%%%%%%%%%%%%%%%%%%%%%%%%%%%%%%%%%%%%%%%%%%%%%

\section{Introduction}
\label{sec:introduction}

% context: startups fail => identification
Startups are ventures undertaken by entrepreneurs to seek, develop, and validate a business model \citep{Katila.2012}. For investors, startups represent investment opportunities with substantial financial risks yet often also with the prospect of large returns. Return on investment can easily exceed the initial investment by several orders of magnitude. As an example, the early-stage investment of Peter Thiel of 0.5 million USD into Facebook increased in value by more than 1 billion USD \citep{CNN.2020}. However, successful investments into startups are rare. Many startups cannot establish an economic business model and eventually fail \citep{Tauscher.2017}. Given the high failure rate among startups, investors are confronted with the non-trivial decision-making task of identifying startups that will eventually be successful \citep{Huang.2015,Scott.2020}.

% VC platforms

In order to find successful startups, investors can nowadays access information about startups through online platforms for venture capital (VC). A prominent example is Crunchbase, where startups can present their venture to investors through a detailed online profile. The online profiles can include both (i)~fundamental variables, which provide structured information on founders, funding, and the business sector, and (ii)~textual self-description. The latter is a free text that can be used to describe the startup in verbal form. Startups can use such online profiles to inform about the venture's prospects and attract the interest of venture capitalists and other potential investors \citep{Connelly.2011,Parhankangas.2014}.

% Prior work
{Prior literature has explored the potential of leveraging VC platform data (e.g., from Crunchbase) to predict startup success due to their comprehensive coverage \citep[e.g.,][]{Alamsyah.2018, Arroyo.2019, Dellermann.2017, Sharchilev.2018, Weibl.2019}. However, prior studies have primarily assessed the predictive power of fundamental variables \citep[e.g.,][]{Alamsyah.2018, Arroyo.2019, Dellermann.2017}, while mostly ignoring textual self-descriptions. Notable exceptions are \citet{Kaiser.2020} and \citet{Sharchilev.2018}, who use textual self-descriptions for prediction. However, these works rely on traditional methods with manual feature engineering. We thus contribute to the existing literature stream with a novel, fused large language model to combine textual self-descriptions with fundamental variables for predicting startup success.}

% online profiles: informative (=> signaling theory)

In this paper, we aim to predict startup success from online profiles of VC platforms. Thereby, we not only consider fundamental information (e.g., on founders, funding, and the business sector) that are captured in traditional scorecards but we also leverage the textual self-descriptions in online profiles on VC platforms. Here, we develop a machine learning approach to predict startup success from large-scale VC platforms. Machine learning allows us to assess how well startup success can be detected for {new} startups and thus support the decision-making of investors regarding whether to select a startup for funding. Specifically, we develop a tailored, \emph{fused} machine learning approach for predicting startup success that considers both (structured) fundamental variables and (unstructured) textual self-descriptions. For this, we draw upon {large language models} as a recent innovation in machine learning \citep{Devlin.2018}, which we carefully adapt to our research objective. A key benefit of large language models in practice is that they are pre-trained on a large amount of public data, because of which relatively small datasets are sufficient for fine-tuning and, thus, to generate accurate predictions. We then assess the relative contribution of textual self-descriptions to making predictions of startup success. 

% findings

We evaluate our machine learning approach based on our fused large language models for predicting startup success using 20,172 online profiles from Crunchbase. Crunchbase is one of the largest online VC platforms hosting online profiles from startups. We find that only fundamental variables can alone make predictions with a balanced accuracy of {72.00}\,\%. When additionally incorporating textual self-descriptions, the balanced accuracy increases to {74.33}\,\%. The improvement is statistically significant, implying that textual self-descriptions are effective in predicting startup success. {In addition, we estimate the financial performance of our machine learning approach by translating the performance improvement to investment portfolio improvement. The investment portfolio improvement amounts to a 40.61 percentage points increase in return on investment (ROI) when incorporating textual self-descriptions, highlighting the practical implications of our machine learning approach. }We then evaluate the prediction performance across various events indicating startup success (\ie, initial public offering, acquisition, and external funding). We further provide an extensive series of sensitivity analyses in which we compare the prediction performance across business sectors, startup age, and additional machine learning baselines, thereby confirming the robustness of our findings.

% implications

Our work contributes to business analytics in several ways. First, we provide empirical evidence on the operational value of online VC platforms for better investment decision-making. Thereby, we extend upon extensive research which has studied the benefits of online platforms for users while we focus on investors. Second, we contribute to a growing stream of machine learning in business analytics \citep[e.g.,][]{Bastani.2021,Choi.2018,Cohen.2018,Misic.2020}. Here, we demonstrate an impactful application of machine learning in VC decision-making. Third, we show the operational value of large language models for research and practice. However, as we detail later, a na{\"i}ve application of large language models would miss significant predictive power. Instead, our task requires a non-trivial adaptation through a \emph{fused} large language model to our decision problem in order to make combined predictions from both fundamental information and texts. Fourth, we provide a flexible tool for investors to automate their screening process in VC decision-making.

% outline

The rest of this paper is structured as follows. \Cref{sec:background} provides a background on venture capital and analytics for decision-making. In \Cref{sec:methods}, we develop our machine learning approach in the form of a tailored large language model. \Cref{sec:data} presents our dataset with online profiles from Crunchbase, based on which we study the predictive power of textual self-descriptions (\Cref{sec:results}). We then discuss implications for both business analytics practice and research (\Cref{sec:discussion}), while \Cref{sec:conclusion} concludes.

\section{Related Work}
\label{sec:background}

\subsection{Venture Capital}

% startups: definition, objective, success

Startups are new entrepreneurial ventures founded to develop and validate a business model \citep{Katila.2012}. In practice, startups typically take an innovative idea and then build a scalable business model around it, with the intention of turning the startup into a high-growth, profitable company \citep{Forbes.2019}. This process is largely dependent on external funding in order to cover costs for technology development, entering markets, or other upfront investments. Hence, events in which startups receive funding are commonly used in the literature to determine success \citep{Arroyo.2019,Hegde.2014,Nanda.2013}. Examples of such events are initial public offerings (\eg, Airbnb, which went public in December 2020), acquisitions (\eg, Slack and DeepMind, which were acquired by Salesforce and Google, respectively), and external funding (\eg, SpaceX, which had several funding rounds after its series A funding in 2002). To capture startup success, prior literature has often studied either individual events such as initial public offerings \citep[\eg][]{Chang.2004,Mann.2007} or a combination of events \citep[\eg][]{Arroyo.2019,Hegde.2014,Nanda.2013}.

% startup: investment side

Startups often represent lucrative investment opportunities with the prospect of large returns. As of 2023, more than 180 startups have turned into \textquote{unicorns,} that is, reached valuations of over USD 1 billion in less than five years \citep{Fleximize.2023}. Investing in such unicorns in an early stage can create a return multiple times larger than the initial investment. However, investments in startups are known to be of high risk. {Startups that eventually fail leave} the investor with little or no return. Hence, identifying successful startups at an early stage is difficult \citep{Aggarwal.2013,Huang.2015,Scott.2020}.

% determinants of success

Predicting which startups will turn out to be successful is inherently challenging, as startups represent new ventures for which little to no information on past performance is available. Thus, many investors make such predictions based on their gut feeling \citep{Huang.2015}. However, according to prior literature, there are several determinants that characterize successful startups. These can be loosely divided into characteristics regarding the business model, the founders, and funding. (1)~The business model explains---to some degree---the survival of ventures \citep{Bohm.2017,Weking.2019}. In this regard, the business sector is also associated with startup success \citep{Holmes.2010,Gelderen.2005}. (2)~Founders decide upon how a business is run and thus founder characteristics are important success factors \citep[\eg,][]{Littunen.2010,Lussier.1995}. For instance, startups are more likely to be on a path toward growth when their founders have attended higher education \citep{Gelderen.2005}. 
 (3)~Funding is often a prerequisite to stimulate growth \cite[\eg,][]{Butler.2020,Conti.2020,Lussier.1995}. Hence, startup success is also associated with previous funding rounds \citep{Baum.2004}. In this regard, it is further beneficial for startups to have backing from a known venture capitalist \citep{Conti.2020,Nahata.2008}. Hence, to avoid relying on gut feeling or subjective bias when processing information about startups, machine learning presents a scalable, data-driven approach to predict startup success.

\subsection{Predicting Startup Success}

% predictions: non-platform

Prior works have developed data-driven approaches for predicting startup success. For instance, predictions can be made based on data from questionnaires, namely via so-called scorecards \citep{Bohm.2017,Yankov.2014}. One study also draws upon data that are extracted from business plan competitions \citep{McKenzie.2017}. Yet, both questionnaires and business plan competitions involve data from manual reporting, which is often not available in VC markets. These data sources also tend to have limited coverage, and thus their usefulness in the daily decision-making of investors is limited. A different stream of literature predicts acquisitions as a specific event in startup lifecycles using proprietary databases (e.g., COMPUSTAT \citep{Ragothaman.2003}, SDC Platinum \citep{Wei.2008}). However, such databases are typically restricted to specific events (\ie, acquisitions) and, on top of that, have limited coverage as they provide only few variables (e.g., about founders and funding) but not textual descriptions. In contrast to that, textual descriptions about startups and their business model, innovation, or market structure may provide significant predictive power regarding which ventures will eventually be successful. 

% predictions: platforms

Recently, the possibility of using online data from VC platforms to predict startup success was explored \citep{Alamsyah.2018,Arroyo.2019,Dellermann.2017,Sharchilev.2018}. Predicting startup success from VC platforms has a clear advantage in practice: online platforms for VC typically exhibit comprehensive coverage of startups and thus provide \textquote{big} data \citep{Weibl.2019}. This is beneficial, as large-scale datasets are generally a prerequisite for making accurate inferences using machine learning. Studies predicting startup success based on questionnaires have often relied on samples with less than 200 observations (\ie, 200 different startups), because of which the prediction models can \emph{not} generalize well across startups and thus lack predictive power \citep{Bohm.2017,Yankov.2014}. In contrast, online platforms for VC, such as Crunchbase, provide online profiles of more than 20,000 startups in the U.\,S alone. 

Based on data from VC platforms, a variety of research questions have been studied. \citet{Arroyo.2019} evaluate the predictive ability of fundamental information at Crunchbase, but textual self-reports as predictors are ignored. In \citet{Dellermann.2017}, a hybrid machine learning approach is designed in which both fundamental information and judgment scores from crowds are combined, but again textual self-reports as predictors are again ignored. \citet{Sharchilev.2018} use textual self-reports for prediction but rely on traditional, bag-of-words representations and not large language models. {\citet{Kaiser.2020} also make predictions from self-reports but rely on a dictionary-based approach that requires manual feature engineering}. Hence, the ability of large language models together with textual self-descriptions from VC platforms has yet to be explored and presents our contribution.

\subsection{Machine Learning in Business Analytics}

\label{sec:NLP}

Machine learning can support managerial decision-making by predicting uncertain operational outcomes \citep{Choi.2018,Cohen.2018,feuerriegel2022bringing,Kraus.2020}. The adoption of machine learning in business analytics has been greatly fueled by the increasing availability of data and recent methodological advances \citep{Bastani.2021, Misic.2020}.  
Promising examples include credit scoring \citep{kriebel2022credit,lessmann2015benchmarking,maldonado2017cost,verbraken2014development}, financial risk assessment \citep{kim2020can,kozodoi2022fairness}, business failure prediction \citep{borchert2023extending,de2020cost,naumzik2022will,stevenson2021value}, throughput prediction \citep{senoner2023addressing}, customer analytics \citep{de2018new, ozyurt2022deep}, recommendation systems \citep{geuens2018framework}, and public sector operations \citep{jakubik2022, kadar2019public}. However, the aforementioned works build upon structured data and not text. 

Business analytics has also increasingly embraced machine learning that can make inferences from textual content \citep[e.g.,][]{borchert2023extending, Cui.2018, feuerriegel2019news, feuerriegel2024generative, haupt2018robust, kraus2017decision,  kriebel2022credit, Lau.2018,   stevenson2021value, toetzke2022monitoring}. As such, business analytics can mine user-generated content, e.g., from social media, in an automated and scalable manner \citep{Cui.2018}. For example, \citet{Cui.2018} enrich historical sales data with social media as a measure of customer perception towards products and evaluate how that combined data source is better in predicting future sales.  However, existing methods in business analytics oftentimes build upon bag-of-words approaches where an unordered set of words is used as input \citep[e.g.,][]{Cui.2018, feuerriegel2019news, Lau.2018} and where, as a result, the relationship, order, and hierarchical structure among words is lost. Hence, existing methods merely operate on word frequencies and not on semantic meaning. A potential remedy is given by large language models that model the ordered sequence of words and thus capture the semantics of running text; however, the operational value of large language models has so far been largely unclear. Moreover, we are not aware of previous work that uses large language models for startup prediction to support investment decisions.

\section{Empirical Model}
\label{sec:methods}

In this section, we first formulate our research question of whether textual self-descriptions from VC platforms predict startup success. To answer this, we then describe our machine learning approach based on a tailored, fused large language model. 

\subsection{Research Question}

% intro

In this study, we build a machine learning approach where we leverage information provided by startups on VC platforms in order to predict startup success. Information on online VC platforms such as Crunchbase can loosely be grouped into two categories (which may potentially complement each other). (1)~VC platforms provide structured information on a startup's fundamentals. Examples of such fundamental variables are the age of the startup, the number of founders, or information about past funding success. Fundamental variables are typically entered on VC platforms in a structured format and thus with little degree of customization. (2)~Startups can additionally provide a textual self-description. The textual self-description can be used to describe the business model, a startup's innovation, or the market structure. Textual self-descriptions have become mandatory on VC platforms such as Crunchbase but the actual content is at the startup's discretion.

In this study, we examine whether large language models can be successfully leveraged by investors to predict startup success from textual self-descriptions on VC. There are several factors that lead us to expect that textual self-descriptions are predictive. In particular, startups can use the textual self-description on VC platforms to present information on a startup's business model, innovation, or market structure. An example is \textquote{BetterTrainers has a new type of business model that protects all sessions booked through the site with premium insurance coverage} where a business model is explained, or \textquote{FaceTec's patented, industry-leading 3D Face Authentication software anchors digital identity with 3D FaceMaps} where a startup details how to make use of certain technologies. Besides the actual information captured in the text, {latent factors such as} the tone of the text (e.g., a positive sentiment) may also implicitly signal success.

As seen by the previous examples, traditional approaches from machine learning for making predictions from online descriptions (e.g., bag-of-words) will likely struggle with the complexity of the underlying task since traditional approaches only rely upon word frequencies and do not provide a principled approach to infer semantic meanings. To this end, the previous examples motivate the use of large language models in our study as a principled, data-driven approach to capture semantic meanings in text and thus to predict startup success.

\subsection{Fused Machine Learning Approach}

\begin{figure}
    \centering
    \includegraphics[width=.85\linewidth]{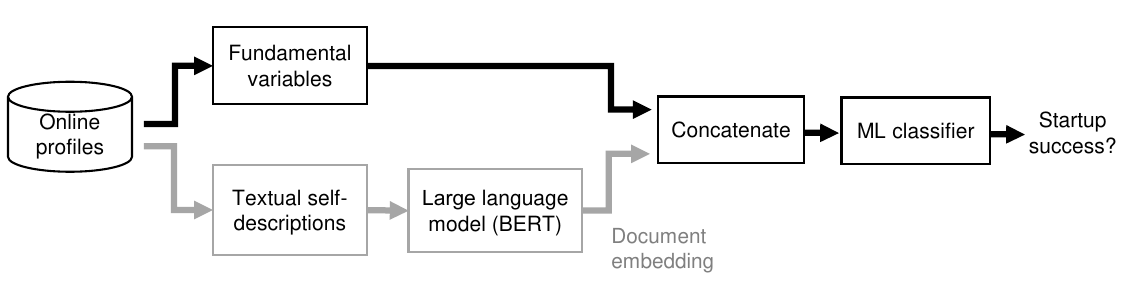} 
    \caption{Our machine learning approach based on a tailored, {fused} large language model for predicting startup success.}
    \label{fig:ml_approach}
\end{figure}

In the following, we present our \emph{fused} machine learning approach in order to predict startup success. Let $i = 1, \ldots, n$ denote the startups. Specifically, we develop a tailored, fused large language model as shown in \Cref{fig:ml_approach}. In our machine learning approach, both sets of variables -- i.e., fundamental variables {(FV)} and textual self-descriptions {(TSD)} -- are taken into account but in different ways. (1)~The fundamental variables come in a structured format $x_i^\text{FV} \in \mathbb{R}^{m_\text{FV}}$ and are thus directly passed on to the final machine learning classifier. (2)~The textual self-descriptions are first mapped onto document embeddings $x_i^\text{TSD} \in \mathbb{R}^{m_\text{TSD}}$ and then passed on to the final ML classifier. Let us denote the final ML classifier by $\phi_\theta : \mathbb{R}^{m_\text{FV} + m_\text{TSD}} \rightarrow \{ 0, 1\}$ with some parameters $\theta$. Here, the output $y_i \in \{ 0, 1 \}$ indicates whether a startup $i = 1, \ldots$ will be successful ($y_i=1$) or not ($y_i=0$). Crucially, a custom architecture for our large language model is necessary in order to fuse both fundamental variables $x_i^\text{FV}$ and document embeddings $x_i^\text{TSD}$ to make predictions. For comparison, we later evaluate a na{\"i}ve large language model without the ``fused'' structure which uses only $x_i^\text{TSD}$ for prediction.

In our machine learning approach, we take the textual self-description of the startup and use a large language model \citep[\ie, BERT; see][]{Devlin.2018} {as an embedding generator} to map text onto a document embedding. The document embedding is then concatenated with the fundamental variables and the resulting concatenated vector is then used as input to the classifier. Large language models represent state-of-the-art techniques for modeling natural language in machine learning \citep{Jurafsky.2020}. A prominent example is BERT, which has been found effective in capturing complex dependencies such as semantics in textual content \citep{Devlin.2018}. In the following, we detail how we fuse data in our large language model.

\subsubsection{Large Language Model (BERT) {as embedding generator}.}

%We make predictions from textual self-descriptions via a large language model. 
Large language models, often also called transformers, are large-scale deep neural networks that are carefully designed to process running text \citep{Jurafsky.2020}. 
The practical benefit of large language models is that they leverage the strength of large-scale deep neural networks and are thus able to capture context, semantics, structure, and meaning \citep{Jurafsky.2020}.

% BERT

A prominent large language model is BERT \citep{Devlin.2018}. BERT was developed by Google AI and stands for {bidirectional encoder representations from transformer}. BERT has been successful in solving various machine learning tasks for natural language. 
In particular, BERT has been shown to be superior to alternative document representations such as bag-of-words. Methodologically, 

Language models such as BERT map running text onto a new representation called embedding \citep{Devlin.2018}. {Formally, each textual input is first transformed into a sequence of tokens $[\mathrm{[CLS]}, w_1, w_2 \ldots]$ based on the predefined vocabulary of BERT, with $\mathrm{[CLS]}$ being used at the beginning of each sequence. Hence, for each textual input, BERT receives a sequence of individual tokens as input where the tokens are represented by vectors $\mathrm{[CLS]}, w_1, w_2, \ldots \in \mathbb{R}^w$.  The vectors are not ``one-hot-encoded'' as traditionally done in simpler models. Instead, BERT uses an embedding layer to convert the sequence of tokens into dense vector representations $e_{\mathrm{[CLS]}}, e_1, e_2, \ldots \in \mathbb{R}^e$ that are lower-dimensional (i.e., the dimensionality $e$ is much smaller than the dimensionality of a typical one-hot encoding, which is computationally more desirable). Next, the token embeddings are fed into a transformer encoder. A transformer encoder is a neural network designed for sequential data that processes the entire input sequence $[e_{\mathrm{[CLS]}}, e_1, e_2, \ldots]$ simultaneously, rather than sequentially. It relies on two key mechanisms: (a)~positional encodings, which add information about the position of each token to retain the order; and (b)~an attention mechanism, which allows the model to weigh the importance of different tokens dynamically. Thereby, a transformer encoder employs a complex, non-linear process to determine how tokens influence one another. The output of the transformer encoder consists of transformed vectors (embeddings) $[o_{\mathrm{[CLS]}}, o_1, o_2, \ldots]$, which can then be used for various tasks. Specifically, the embedding for the $\mathrm{[CLS]}$ token (i.e., $o_{\mathrm{[CLS]}}$) can be used for classification tasks as it aggregates the meaning of the entire input sequence.} 

\begin{figure}
    \centering
    \includegraphics[width=.4\linewidth]{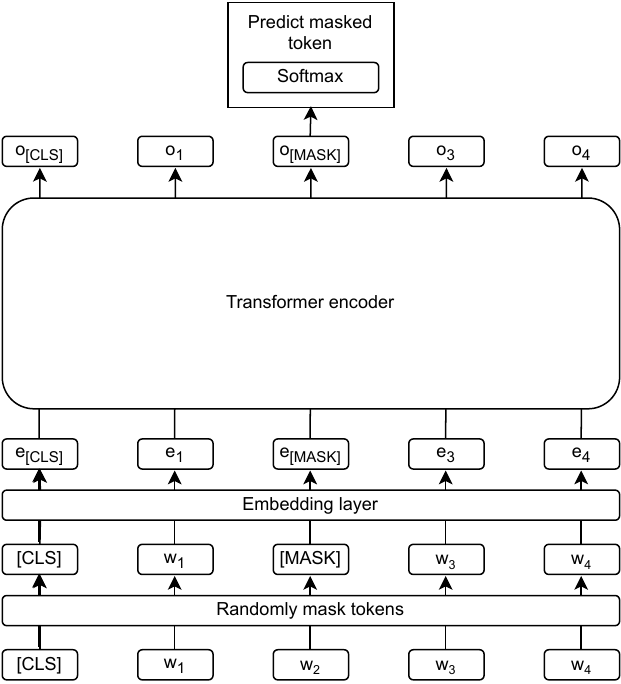} 
    \caption{Schematic overview of a large language model (here: BERT).}
    \label{fig:BERT}
    \vspace{-0.5cm}
\end{figure}

{During training, BERT utilizes a technique called masked language modeling, where some of the input tokens are randomly masked (i.e., omitted) for self-supervised learning. The objective of BERT during training is to correctly predict these masked tokens. Thereby, BERT updates its internal weights and learns a deep understanding of language context and relationships between words.} Due to self-supervised learning, large-scale textual databases (e.g., Wikipedia) can be used for training but without the need for explicitly annotated labels. A schematic visualization is in \Cref{fig:BERT}.

% our implementation

Our implementation is as follows. We use the so-called basic, uncased version of BERT \citep{Devlin.2018}, comprised of 12 layers with $\sim$\,110 million trainable parameters. It generates embeddings $o_{\mathrm{[CLS]}}, o_1, o_2, \ldots \in \mathbb{R}^n$ of dimension $n = 768$. BERT is shipped as a pre-trained network where parameters have already been learned from open-source content. {Before applying BERT, all text is lowercased and tokenized using the WordPiece algorithm, which maps the text onto subwords or unigrams from the WordPiece vocabulary}. Afterward, the text is passed through the pre-trained BERT network. {The embedding of the $\mathrm{[CLS]}$ token (i.e., $o_{\mathrm{[CLS]}}$) is then used as the document embedding $x^\text{TSD}$, representing the textual self-description for the downstream classification. Hence, our document embedding $x^\text{TSD}$ is of dimension $m_\text{TSD}=786$.}

\subsubsection{{Baseline text representations.}}

{We compare our machine learning approach based on a tailored, fused large language model against three traditional text representations. {All of the baselines are again concatenated to the fundamental variables and are then fed into a final machine learning classifier. } The final machine learning classifier is again subject to rigorous hyperparameter tuning (see later for details) for fair comparison. Therefore, all performance gains from our approach must be attributed to that large language models are better at handling textual content. 

\begin{itemize}
\item \textbf{Manual feature extraction: }The first text-based baseline is based on manual feature extraction. Specifically, we manually craft features that capture textual information (e.g., the length, the mean word length, and the number of geographic references). We follow prior literature and extract the same features as in \citet{Kaiser.2020}. {This results in a text representation of dimension 10.} We refer readers to \citet{Kaiser.2020} for a full list of the features.

\item \textbf{Bag-of-words: } We compare our machine learning approach against the traditional approach of a bag-of-words baseline. We refer readers to \citet{Jurafsky.2020} for an introduction. We implement bag-of-words as follows. We first tokenize the words of the textual self-description to unigrams, remove stop words, lemmatize, and apply a tf-idf weighting. Furthermore, we remove words with more than 95\,\% sparsity. {The bag-of-words baseline results in a 98-dimensional text representation.}

\item \textbf{GloVe: } GloVe  \citep{Pennington.2014} transforms words into vectors (so-called word embeddings) based on their co-occurrence in a text corpus. Thereby, the vectors capture semantic relationships, offering a rich set of features for text analysis. We use the GloVe model pre-trained on Wikipedia (i.e., \texttt{glove-wiki-gigaword-50}) to extract the 50-dimensional word embeddings. We average the individual word embeddings to get the final text representation used for the downstream classification.
\end{itemize}
}

\subsubsection{Final Machine Learning Classifier.}

The final machine learning classifier $\phi_\theta(\cdot)$ {with parameters $\theta$} is responsible for the ``fused'' approach and, for this, receives the concatenated vector of (1)~fundamental variables and (2)~document embeddings. The output is then the predicted probability of startup success.  
We thus optimize 
\begin{equation}
\theta^\ast = \argmin_\theta \mathbb{E}[ \mathcal{L}(\phi_\theta([x^\text{FV}, x^\text{TSV}]), y) ] ,
\end{equation}
where $\mathcal{L}$ is a convex loss (e.g., mean squared error) and where $[ \cdot , \cdot]$ is the concatenation operator.

{We experiment with different classifiers that are designed to handle both linear and non-linear relationships in the data. Specifically, we make use of the following classifiers:}

\begin{itemize}
\item \textbf{Logistic regression:} {The logistic regression is a simple linear model used for binary classification. It models the probability of a binary outcome using the logit function to map predictions to probabilities. The logistic model expresses the log-odds of the outcome variable as a linear combination of the independent variables, formalized as $\log \left(\frac{p}{1-p}\right)=\theta^T x$, where p is the probability of the outcome of interest.}
\item \textbf{Elastic net:} The elastic net {extends the logistic regression} in which overfitting is prevented through regularization \citep{Zou.2005}. Specifically, regularization is given by a combination of both an L1- and an L2-norm penalty (analogous to lasso and ridge methods, respectively). This thus shrinks some coefficients closer to zero, and, as a result, the classifier generalizes better to out-of-sample observations. Formally, let $\phi_\theta(x) = \theta^T x$. Then the regularized loss $\mathcal{L}_{\mathrm{reg}}$ is formalized as $\mathcal{L}_{\mathrm{reg}}(x,y) = \mathcal{L}(\phi_\theta(x), y) + \lambda \, \big( \frac{1-\alpha}{2} \norm{\theta}^2_2 + \alpha \, \norm{\theta}_1 \big)$ with hyperparameters $\alpha$ and $\lambda$. The elastic net is especially beneficial in tasks where predictors are subject to linear dependence \citep{Hastie.2009}. For reasons of completeness, we also experimented with lasso and ridge methods \citep{Hastie.2009}, but with qualitatively similar results (and thus omitted the results for brevity). 
\item \textbf{Random forest:} The random forest is an ensemble learning classifier where predictions are made from a multitude of decision trees \citep{Ho.1995}. Each decision tree is fit to a random subset of the data, while the final prediction is then made by taking the majority vote over the individual decision trees. As a result, the classifier is less prone to overfitting, has a better prediction performance than a single decision tree, and is effective in handling non-linear relationships \citep{Hastie.2009}. 
{\item \textbf{Neural net:} The neural network is a flexible model for classification by using layers of nodes that transform the input through non-linear activation functions. The output layer uses a sigmoid to produce class probabilities. The loss is regularized by a combination of both the L1- and L2-norm penalty to prevent overfitting. Neural networks excel due to their flexibility in handling non-linear
relationships.}
\end{itemize}

\subsection{Performance Metrics}

To evaluate the performance of machine learning in predicting startup success, we report different performance metrics: balanced accuracy, precision, recall, $F_1$-score, area under the curve from the precision-recall curve (AUCPR), and area under the curve from the receiver operating characteristics (AUROC). However, due to its inherent benefit of considering the complete distribution of discrimination thresholds \citep{Hastie.2009}, we primarily focus on the AUROC. We remind that we follow common practice in machine learning and evaluate the performance on {out-of-sample observations, that is, startups that have not been part of the training set but from the test set so that they are thus {unseen} to the machine learning classifiers}.

{
Furthermore, we calculate the return on investment (ROI) for the machine learning-selected portfolios. Let $\mathit{TP}$ denote the number of true positives (\ie, cases where startup success was predicted correctly) and $\mathit{FP}$ the number of false positives (\ie, cases where the model predicted success despite that the startup was actually not successful). We then calculate the net investment gain for correctly predicted successful startups by taking the sum of the final investment values $\mathit{FIV_{TP}}$ (\ie, the startup valuations after a success event) minus the sum of the total cost of investment ($IC$).
Note that, since data on startup valuations and costs of investment is not always publicly available, we approximate these variables using constants determined based on historical mean values for startups listed on Crunchbase.\footnote{\SingleSpacedXI\footnotesize In our Crunchbase dataset (see \Cref{sec:data}), the valuation of a startup after a success event (\ie, initial public offerings, funding, acquisitions) is, on average, \$184.47 million. The pre-success valuation (\ie, the last valuation in previous funding rounds) is, on average, \$12.19 million. Hence, startups have, on average, a 15.13 times higher valuation if they become successful.}
For companies that were non-successful, %during the simulation period, 
we conservatively assign a final investment value of zero ($\mathit{FIV_{FP}}$). The ROI for the portfolio is then calculated by taking the net investment gain divided by the total cost of investment. Formally,
\begin{equation} \label{eqn:roi}
	ROI = \frac{\mathit{TP} \times \mathit{FIV_{TP}} + \mathit{FP} \times \mathit{FIV_{FP}} - (\mathit{TP} + \mathit{FP}) \times \mathit{IC}}{(\mathit{TP} + \mathit{FP}) \times \mathit{IC}} \times 100,
\end{equation}
where the total investment costs ($\mathit{IC})$ comprise (i) the investor's investment into equity of the startup; and %last valuation of a startup prior to the investment; (2) %the simulation period (mean of \$12.19 million) 
(ii) we consider 10\% of the last valuation as additional screening and monitoring costs for the investor.
}
   
\subsection{Implementation Details}

%(Prediction system)

{Our implementation follows best practice in machine learning \citep{Hastie.2009}. For this, we split the data into a training set and a test set. The former is used for training the model; the latter is used to evaluate the out-of-sample performance. In our work, we randomly assign 80\,\% of the data samples to the training set and 20\,\% to the test set.} Due to class imbalances, common procedures in machine learning are followed; that is, we apply a stratified split \citep{Goodfellow.2016}, so that both sets have the same ratio of successful vs. non-successful startups. {To ensure robustness in our evaluation, we repeat the random split five times and report the mean and standard deviation of the performance metrics on the test set across the five iterations}. This allows us to quantify how well machine learning can predict success for ventures that were not seen during training. 

Hyperparameter tuning is conducted using 10-fold cross-validation. Specifically, hyperparameters are tuned via {randomized} grid search {(20 iterations)}, using the tuning grid in \Cref{tbl:tuning_grid}. {The best hyperparameters are selected based on the cross-validated AUROC score}. Note that the hyperparameter tuning is done separately for the different input variables, that is, for when training our machine learning approach using fundamental variables (FV), textual self-descriptions (TSD), or a combination of both (FV + TSD).

\begin{table}
\TABLE
{Tuning grid for hyperparameter tuning.\label{tbl:tuning_grid}}
{
\OneAndAHalfSpacedXI
\tiny
\begin{tabular}{lllr}
\toprule
{\textbf{Classifier}} & {\textbf{Hyperparameter}} & {\textbf{Tuning range}}\\% & 

 & & \\
\midrule 
  {Logistic regression} & {---} & \\

\midrule 
  Elastic net & {L1 ratio} & \{{0.1, 0.2, 0.3, 0.5}\}\\
  & {Tolerance for stopping} & \{{0.00001, 0.0001, 0.001, 0.01}\}\\
   & Regularization parameter & \{{0.1, 0.25, 0.5, 1, 2, 4, 8}\}\\
\midrule 
  Random forest & {\% of randomly selected predictors} & \{{0.4, 0.6, 0.8}\}\\
   & Splitting rule & \{gini, {entropy}\}\\
   & Minimal node size & \{{5, 8, 10}\}\\
   & {Minimal split size} & \{{8, 10}\}\\
\midrule 
  Neural network & \# Hidden units & {$\mathit{input\_dim}*$\{{0.25, 0.5, 0.75, 1, 1.5, 2}\}}\\
	 & {\# Hidden layers} & {2, 3, 4}\\
   & Dropout rate & \{0, 0.2\}\\
   & Batch size & \{{128, 256, 512}\}\\
   & {Optimizer} & {AdamW \citep{Loshchilov.2018}} \\ 
   & Learning rate & \{0.001{, 0.0001, 0.00001}\}\\
   & {Max Epochs} & {500} \\ 
   & Learning rate decay & 0\\
   & Activation function & ReLU\\
   & {Early stopping patience} & {3} \\ 
\bottomrule
\end{tabular}
}
{
\hspace{-0.5cm} {\emph{Note:} $\mathit{input\_dim}$ is the number of input features and therefore depends on whether the fundamental variables, the textual self-description, or a combination of both is used for training.}
\vspace{-.5cm}
}
\end{table}

\section{Empirical Setting}
\label{sec:data}

\subsection{Online Profiles on Crunchbase}

% Crunchbase

Our evaluations are based on data from Crunchbase.\footnote{\url{http://www.crunchbase.com}} Crunchbase is a leading online VC platform that connects startups and investors. For this, Crunchbase allows startups to create online profiles where they can present information on their business, founders, and funding. Edits can be made by verified employees to ensure that correct information is entered.

We collected online profiles (\ie, both fundamental variables and textual self-descriptions) from all US-based startups that were listed on Crunchbase. Furthermore, we excluded startups that went public and that have already received series C funding (or a later funding round). The latter is important as our objective is to make predictions for companies that fall under the definition of a startup.

\subsection{Definition of Startup Success}

In our study, we predict startup success with regard to different events that are conventionally used as indicators of success \citep[cf.][]{Arroyo.2019,Hegde.2014,Nanda.2013}, namely, whether startups had an initial public offering, have been acquired, or secured external funding. If any of these events occurred, we treat the startup as \textquote{successful.} Otherwise, a startup is treated as \textquote{non-successful.} If not stated otherwise, these labels are used to evaluate our machine learning approach. As part of our sensitivity analyses, we later continue to compare how the prediction performance varies across these events -- \ie, initial public offerings, funding, and acquisitions.

\subsection{Time-Aware Prediction and Evaluation Framework}

We implemented a \emph{time-aware} approach that is common in time-series forecasting \citep{Hastie.2009}. Recall that we aim to evaluate whether we can predict if startups will become successful in the {future}. Consequently, we processed our data as follows. We restricted our analysis to startups that were founded between 2013 and 2015, based on which we predicted their future development until the end of 2020. We obtained raw access to the Crunchbase database with historical data. This allowed us to collect information from online profiles that were available in 2015. In particular, we discarded information that was added or updated later, so that we only considered data as presented on Crunchbase at the end of 2015. 

We then predict whether an event indicating startup success has occurred during the years 2016 through 2020, that is, we make forecasts whether startups were successful over a time horizon of five years. The forecast horizon is set analogous to earlier statistics reporting upon a high failure rate among startups in their early stage \citep{LaborStatistics.2016}, so that a 5-year-ahead forecast horizon should be sufficient to distinguish successful from non-successful startups. Our choice of events representing startup success is listed in the previous section.

\subsection{Variable Descriptions}

% dependent variable

Our fused machine learning approach makes use of an extensive set of variables from Crunchbase (see \Cref{tbl:variables}). The outcome variable (\ie, the variable to predict) is binary, denoting whether a startup was successful ($=1$; otherwise $=0$). 

\begin{table}[htb]
\TABLE
{Variable descriptions.\label{tbl:variables}} 
{
\OneAndAHalfSpacedXI
\tiny
\scriptsize
\begin{tabular}{p{5cm}p{11cm}}
\toprule
\textbf{Variable} & \textbf{Description}  \\
\midrule
			\multicolumn{2}{c}{\textsc{Outcome variable}} \\
			\midrule
  Success & True ($=1)$ if a startup had an initial public offering, received funding, or has been acquired. False ($=0$) otherwise.\\ 
				\midrule
			\multicolumn{2}{c}{\textsc{Predictors}} \\
			\midrule

\multicolumn{2}{l}{\underline{Fundamental variables (FV)}} \\
  Age & Time since the startup has been founded (in months) \\ 
  \addlinespace
  Has email & Whether the startup has added an email address ($=1$ if true, otherwise 0) \\ 
  Has phone & Whether the startup has added a phone number ($=1$ if true, otherwise 0) \\ 
  Has Facebook & Whether the startup has added a link to Facebook ($=1$ if true, otherwise 0) \\ 
  Has Twitter & Whether the startup has added a link to Twitter ($=1$ if true, otherwise 0) \\ 
  Has LinkedIn & Whether the startup has added a link to LinkedIn ($=1$ if true, otherwise 0) \\ 
  
  \addlinespace
  Founders count & Number of founders of the startup \\ 
  Founders country count & Number of unique countries the founders are from \\ 
  Founders male count & Number of male founders \\ 
  Founders female count & Number of female founders \\ 
  Founders degree count total & Total number of university degrees of the founders \\ 
  Founders degree count maximum & Number of degrees for most educated founder \\ 
  Founders degree count mean & Average number of degrees per founder \\ 
  
  \addlinespace
  Number of investment rounds & Number of investment rounds \\ 
  Raised funding & Total raised funding (in million USD)\\ 
  Last round investment type & Investment type (seed, series A, etc.) of the last funding round \\
  Last round raised funding & Raised funding (in million USD) in the last funding round \\ 
  Last round post money evaluation & Valuation (in million USD) of the startup after the last funding round \\ 
  Last round time lag & Time since last funding round (in months) \\ 
  Investor count & Overall number of investors that invested in the startup \\ 
  Last round investor count & Number of investors in the last funding round only \\ 
  Known investor count & Overall number of investors with a profile on Crunchbase \\ 
  Last round known investor count & Number of investors in the last funding round with a profile on Crunchbase \\ 
	
	\addlinespace
	Industries & Fine-grained industries in which the startup operates (according to the Crunchbase coding scheme; e.g., ``machine learning'', ``machinery manufacturing'') \\

  \addlinespace
\multicolumn{2}{l}{\underline{Textual self-description (TSD)}} \\
  Document embedding & Textual self-description encoded via large language model (BERT) \\
  
\bottomrule

\end{tabular}
}
{
\vspace{-.5cm}
}
\end{table}

% predictors

The predictors (\ie, the variables that are fed into our machine learning classifiers) consist of the following: (structured) fundamental variables and (unstructured) textual self-descriptions. (1)~The fundamental variables (FV) describe different characteristics of startups such as their age or the industries in which they operate (see \Cref{tbl:variables}). Note that we use the industries as reported on Crunchbase, which is based on a highly granular scheme (e.g., an Internet-of-Things company may be assigned simultaneously to ``artificial intelligence'', ``industrial automation'', etc.). Social media activity has been found to be related to startup success \citep{Jin.2017}, and, analogously, we include information about whether startups are on social media (\eg, whether they have a Twitter/X or LinkedIn profile). Furthermore, we collect information about the characteristics of the founders (\eg, the number of university degrees). We follow previous literature \citep{Conti.2020} by controlling for the presence of known investors that have a profile on Crunchbase themselves. We also include information on previous funding rounds but, since we use a historical view on Crunchbase data, we only access information up to our time point when making the predictions so that there is \emph{no} lookahead bias (i.e., we discard funding rounds that occur during the forecast horizon to ensure a time-aware evaluation framework).\footnote{\SingleSpacedXI\footnotesize 
We also considered a less sparse encoding of business sectors (rather than fine-grained industries as in the Crunchbase coding scheme) but we discarded this. The reason is seen in our later analysis, where there is little variability across business sectors and, thus, sector information has only little predictive power. Furthermore, we also considered additional information about founders (e.g., their number of current and past jobs) and prominence (e.g., site visitors, growth in site visitors, number of media articles) but found that these are too sparse to make a meaningful addition to our predictions.} (2)~The latter, \ie, textual self-descriptions (TSD), are encoded via the large language model (BERT). This yields document embeddings, which are then used as input to the machine learning classifier.

\subsection{Descriptive Statistics}\label{sec:descriptives}

The above filtering yields a final dataset with \num{20172} startups. Descriptive statistics on startups for our dataset are as follows (see \Cref{tbl:descriptives}). Out of all startups, \num{7252} (\ie, 35.94\,\%) startups have been labeled as successful, whereas \num{12920} (\ie, 64.06\,\%) have been labeled as being non-successful. Frequent events indicating success are founding rounds (\ie, 32.45\,\% of all startups), followed by acquisitions (3.10\,\%) and initial public offerings (0.40\,\%). For startups in our dataset, the average age is 18 months. Startups tend to be more successful if they provide a link to their social media profiles. In general, startups are more frequently founded by males (\ie, 1.58 male founders per startup) than by females (\ie, 0.25 female founders per startup). Successful startups have, on average, more founders (mean: 1.98) than non-successful ones (mean: 1.69). Furthermore, founders with university degrees often have more successful startups. On average, startups have previously raised funding totaling to USD \num{3.156} million from 2.07 investors. Unsurprisingly, startups that are eventually labeled as successful have received more funding (mean: USD 4.81 million) and are backed by more investors (mean: 3.28). {On average, successful startups provide a shorter textual self-description (mean: $613.32$ characters) than non-successful ones (mean: $694.04$ characters). \Cref{tbl:tsdexamples} lists two example textual self-descriptions, one for each class.}

\begin{table}[H]
\TABLE
{Descriptive statistics.\label{tbl:descriptives}}
{
\OneAndAHalfSpacedXI
\tiny
\sisetup{round-mode=places,round-precision=2}
\begin{tabular}{lS[table-format=2.4]S[table-format=2.4]S[table-format=2.4]S[table-format=2.4]S[table-format=2.4]S[table-format=2.4]}
  \toprule
\textbf{Variable} & \multicolumn{2}{c}{\textbf{Overall}} & \multicolumn{2}{c}{\textbf{Non-successful}} & \multicolumn{2}{c}{\textbf{Successful}} \\
\cmidrule(lr){2-3} \cmidrule(lr){4-5} \cmidrule(lr){6-7}
& {\textbf{Mean}} & {\textbf{SD}} & {\textbf{Mean}} & {\textbf{SD}} & {\textbf{Mean}} & {\textbf{SD}} \\ 
  \midrule
  
			\multicolumn{7}{c}{\textsc{Outcome variable}} \\
			\midrule
  Success & 0.36 & 0.48 & 0.00 & 0.00 & 1.00 & 0.00 \\   
				\midrule
			\multicolumn{7}{c}{\textsc{{Fundamental variables (FV)}}} \\
			\midrule  
  Age (in months) & 18.14 & 10.01 & 19.54 & 9.85 & 15.66 & 9.81 \\ 
	\addlinespace
  Has email & 0.77 & 0.42 & 0.75 & 0.43 & 0.82 & 0.38 \\ 
  Has phone & 0.58 & 0.49 & 0.58 & 0.49 & 0.58 & 0.49 \\ 
  Has Facebook & 0.73 & 0.45 & 0.73 & 0.44 & 0.72 & 0.45 \\ 
  Has Twitter & 0.77 & 0.42 & 0.75 & 0.43 & 0.79 & 0.41 \\ 
  Has LinkedIn & 0.70 & 0.46 & 0.62 & 0.49 & 0.84 & 0.36 \\ 
  \addlinespace
  Founders count & 1.83 & 0.97 & 1.69 & 0.89 & 1.98 & 1.03 \\ 
  Founders different country count & 1.20 & 0.42 & 1.17 & 0.39 & 1.23 & 0.45 \\ 
  Founders male count & 1.58 & 1.03 & 1.43 & 0.94 & 1.74 & 1.08 \\ 
  Founders female count & 0.25 & 0.52 & 0.26 & 0.52 & 0.24 & 0.51 \\ 
  Founders degree count total & 1.16 & 1.46 & 0.88 & 1.22 & 1.46 & 1.62 \\ 
  Founders degree count maximum & 0.84 & 0.92 & 0.69 & 0.88 & 1.00 & 0.94 \\ 
  Founders degree count mean & 0.78 & 0.84 & 0.66 & 0.83 & 0.92 & 0.83 \\   
    \addlinespace
  Number of investment rounds & 1.36 & 0.73 & 1.25 & 0.61 & 1.51 & 0.83 \\ 
  Raised funding (in million USD) & 3.16 & 21.45 & 1.83 & 21.58 & 4.81 & 21.17 \\ 
  Last round raised funding (in million USD) & 2.59 & 22.23 & 1.73 & 22.92 & 3.85 & 21.12 \\ 
  Last round post money evaluation (in million USD) & 12.19 & 48.39 & 8.48 & 36.06 & 16.84 & 60.19 \\ 
  Last round time lag (in months) & 11.59 & 8.55 & 14.14 & 8.80 & 8.40 & 7.03 \\ 
  Investor count & 2.07 & 3.82 & 1.11 & 2.55 & 3.28 & 4.70 \\ 
  Last round investor count & 1.42 & 2.65 & 0.79 & 1.88 & 2.34 & 3.27 \\ 
  Known investor count & 1.16 & 0.68 & 1.06 & 0.38 & 1.30 & 0.91 \\ 
  Last round known investor count & 1.08 & 0.46 & 1.02 & 0.25 & 1.16 & 0.62 \\
    \midrule
			\multicolumn{7}{c}{\textsc{{Textual self-descriptions (TSD)}}} \\
			\midrule  
   {TSD length in chars (only for descriptive purpose)} & {665.02} & {429.26} & {694.04} & {462.13} & {613.32} & {357.64} \\
   \bottomrule
   \multicolumn{3}{l}{SD = standard deviation} &  \multicolumn{4}{r}{$N$ = 20,172 startups}
\end{tabular}
}
{
}
\vspace{-1cm}
\end{table}

\begin{table}[H]
\TABLE
{Examplary textual self-descriptions for each class.\label{tbl:tsdexamples}}
{
\OneAndAHalfSpacedXI
\tiny
\renewcommand{\arraystretch}{1.5}
\sisetup{round-mode=places,round-precision=2}
{
\begin{tabular}{p{13.cm}p{2.cm}}
\toprule                                  {Textual self-description} &                                         {Outcome} \\
\midrule
``\texttt{Fluc (which is Miles \& Company Services now) is building the world’s first social marketplace for consumers to search, discover, and purchase freshly-made food. Whether people desire a cup of coffee or a freshly cooked pacific trout, Fluc powers the connection between consumers and local food merchants.  Fluc wraps complex logistics into a simple and affordable consumer experience, enabling anyone to access thousands of food items from the palm of their hand.}''  & non-successful \\
``\texttt{Lemonade is a licensed insurance carrier that offers homeowners and renters insurance powered by artificial intelligence and behavioral economics.  By replacing brokers and bureaucracy with bots and machine learning, Lemonade promises zero paperwork and instant everything.  And as a Certified B-Corp, where underwriting profits go to nonprofits, Lemonade is remaking insurance as a social good, rather than a necessary evil.}'' & successful\\
\bottomrule
\end{tabular}
}
}
{
}
\end{table}

% business sectors

Startups listed on Crunchbase operate in a variety of business sectors (see \Cref{tbl:sector_allocation}). The majority of startups in our data operate in the area of \textsc{Information Technology} and \textsc{Communication Services}. In contrast, startups in the \textsc{Energy}, \textsc{Utilities}, and \textsc{Materials} sectors are less common. Note that startups can be assigned to multiple business sectors. Across the business sectors, we also see variation in the success rate of startups. For instance, startups in some sectors such as \textsc{Utilities} have a high success rate (50.79\,\%), while the success rate for \textsc{Communication Services} amounts to only 32.71\,\%.

\begin{table}[H]
\TABLE
{Relative frequencies and success rates of startups across different business sectors.\label{tbl:sector_allocation}}
{
\OneAndAHalfSpacedXI
\tiny
\sisetup{round-mode=places,round-precision=2}
\begin{tabular}{lS[table-format=2.2]S[table-format=2.2]}
  \toprule
{\textbf{Business sector}} & {\bfseries\mcellt{Relative\\ freq. (in\,\%)}} & {\bfseries\mcellt{Success rate\\ (in\,\%)}} \\
  \midrule
  \textsc{Information Technology} & 54.32 & 40.61 \\ 
  \textsc{Communication Services} & 43.23 & 32.71 \\ 
  \textsc{Industrials} & 34.72 & 41.53 \\ 
  \textsc{Consumer Discretionary} & 34.14 & 32.95 \\ 
  \textsc{Health Care} & 16.21 & 52.06 \\ 
  \textsc{Consumer Staples} & 15.86 & 42.06 \\ 
  \textsc{Financials} & 10.01 & 41.14 \\ 
  \textsc{Real Estate} & 5.99 & 35.84 \\ 
  \textsc{Utilities} & 2.21 & 50.79 \\ 
  \textsc{Energy} & 2.16 & 44.14 \\ 
  \textsc{Materials} & 1.52 & 43.97 \\ 
   \bottomrule
\end{tabular}
}
{\hspace{-0.5cm} \emph{Note:} Business sectors are categorized according to the Global Industry Classification Standard (GICS). Startups can belong to multiple business sectors.
\vspace{-.5cm}
}
\end{table}

\section{Empirical Findings}
\label{sec:results}

\subsection{Comparison of Our Large Language Model Against the Baselines}\label{sec:results_basic}

We now evaluate the performance of our fused large language model in predicting startup success (see \Cref{tbl:prediction_bow}). {We use the neural net as the best-performing final machine learning classifier within our fused large language model for this evaluation. For a detailed comparison across different final machine learning classifiers, we refer to \Cref{appendix:long_eval} of the Supplementary Materials.} We further remind that we follow common practice in machine learning and evaluate the performance on out-of-sample observations, that is, startups that have not been part of the training set and are thus unseen to the machine learning classifiers. In addition, we repeat the random splitting of our train and test sets five times and thus report the mean and standard deviation of our evaluation metrics across the five test sets.\footnote{{We also perform an out-of-time evaluation in \Cref{appendix:outoftime_eval} of the Supplementary Materials, where we evaluate the performance in predicting the success of startups that originate from a period outside the one used for training. Overall, the performance remains robust but, due to the task formalization, has a smaller sample size and thus tends to have a larger variance.}}

Overall, we find that our tailored, fused large languages model is considerably more accurate than a majority vote (i.e., a model that always predicts the majority class label) and a random vote (i.e., a model that predicts class labels randomly based on the distribution of the class labels in the training data). Both approaches represent na{\"i}ve baselines from machine learning, which are outperformed by a large margin. Our tailored, fused large language model using both fundamentals and textual self-descriptions yields an AUROC of {82.78}\,\%, a balanced accuracy of {74.33}\,\%, and a {7.23}-fold ROI. Altogether, this demonstrates the efficacy of machine learning based on our fused large language model in predicting startup success from VC platforms.

{We further compare our fused large language model against common baseline {text representations}. Specifically, we draw upon manual feature extraction from textual data \citep{Kaiser.2020}, GloVe document embeddings \citep{Pennington.2014}, and a bag-of-words approach \citep{Jurafsky.2020}. The baseline {text representations} have a known limitation in that they struggle with capturing long-term dependencies across language, because of which semantics are ignored to a large extent. As expected, we find that, compared to our fused large language model, the baselines are inferior. For example, the best baseline in terms of AUROC (GloVe) has a {6.41-fold} ROI, while our custom, fused large language model has a {7.23-fold} ROI, which is a plus of {82.19} percentage points. Note that both our fused, large language model and the bag-of-words baseline have access to the same data, that is, fundamental variables and textual self-descriptions. Hence, all performance improvements must solely be attributed to the better model architecture of our fused large language model.}

\begin{table}[H]
	\TABLE 
	{Prediction performance of our large language model vs. the baselines. \label{tbl:prediction_bow}}
	{
	\OneAndAHalfSpacedXI
	\tiny
        \renewcommand{\arraystretch}{1.2}
	\sisetup{round-mode=places,round-precision=2,detect-weight,mode=text,table-column-width=1.2cm}
 {
	\begin{tabular}{l>{\centering\arraybackslash}p{1.22cm}>{\centering\arraybackslash}p{1.15cm}>{\centering\arraybackslash}p{1.15cm}>{\centering\arraybackslash}p{1.15cm}>{\centering\arraybackslash}p{1.15cm}>{\centering\arraybackslash}p{1.15cm}>{\centering\arraybackslash}p{1.2cm}}
	  \toprule
	{\textbf{Approach}} & {\textbf{Balanced accuracy}} & {\textbf{Precision}} & {\textbf{Recall}}  & {\textbf{$\boldsymbol{F_1}$-score}} & {\textbf{AUROC}} &{\textbf{AUCPR}} & {\textbf{ROI}}  \\ 
	  \midrule
	Majority vote & 50.00 & {\textemdash{$^\dagger$}} & 0.00 & {\textemdash{$^\dagger$}} & 50.00 & 0.00 & {\textemdash{$^\dagger$}} \\  
 Random vote & 50.00 & 36.59 & 36.14 & 36.36 & 50.00 & 23.16 & 403.40 \\  
	\midrule 
  FV only & 72.00 (1.33) & 56.03 (3.27) & 79.46 (4.45) & 65.56 (0.92) &  80.60 (0.44) & 70.92 (0.68) & 670.84 (45.05) \\
  \addlinespace
  FV + Manual feature extraction & 72.40 (1.15) & 56.83 (3.13) & 78.86 (4.65) & 65.89 (0.77) & 81.22 (0.39) & 71.42 (0.63) & 681.88 (43.12) \\
  \addlinespace
  FV + GloVe & 71.87 (1.93) & 53.86 (3.27) & \textbf{85.16} (3.72) & 65.85 (1.48) & 81.89 (0.59) & 72.59 (0.59) & 640.90 (44.93) \\
  \addlinespace
  FV + Bag-of-words  & 72.38 (0.49) & 55.71 (1.23) & 80.95 (1.49) & 65.98 (0.38) &  81.10 (0.18) & 71.74 (0.63) & 666.38 (16.88) \\
  \midrule 
  Our fused large language model (FV + TSD) & \textbf{74.33} (0.25) & \textbf{59.83} (1.79) & 78.28 (2.63) & \textbf{67.77} (0.15) & \textbf{82.78} (0.25) &  \textbf{73.70} (0.49) & \textbf{723.09} (24.56) \\
	 \bottomrule
  \multicolumn{7}{l}{$^\dagger$Value not defined due to division by zero (\ie, there is no successful class)} \\
 \multicolumn{5}{l}{FV = fundamental variables}\\
	\end{tabular}
	}}
	{\hspace{-0.5cm} {\emph{Note:} Reported is the mean (and standard deviation) out-of-sample prediction performance across 5 random splits (in \%). The best value per metric and model is highlighted in bold. K\&K is short for the features from \citet{Kaiser.2020}.}
 \vspace{-.5cm}
 }
 
\end{table}

In addition, we compare using fundamental variables {only} vs. a combination of fundamental variables and {the textual self-description}. Here, including textual self-description {using the baseline text representations} increases the AUROC by {0.62} percentage points ({manual feature extraction \citep{Kaiser.2020}}), {1.29} percentage points ({GloVe \citep{Pennington.2014}}), and {0.5} percentage points ({bag-of-words}). {Including textual self-descriptions within our fused large language model performs best and increases the AUROC by 2.18 percentage points}. As such, we yield consistent evidence that demonstrates the operational value of textual self-descriptions: a significant improvement in prediction performance is achieved by including textual self-descriptions. Altogether, this highlights the importance of textual self-descriptions for successful investing decisions.

\subsection{Sensitivity to Final Machine Learning Classifier}\label{sec:results_sensitivity}
% overview

We now provide a sensitivity analysis where we vary the final machine learning classifier (i.e., elastic net, random forest, neural network) and  {within our fused large language model}.\footnote{{We also tested the performance of varying the input variables (i.e., FV, TSD, FV~$+$~TSD) within our final machine learning classifiers. We report the results in \Cref{apx:input_vars}.}} Thereby, we confirm that our choice of a neural network for the final machine learning classifier in our fused large language model is superior. The results are reported in \Cref{tbl:prediction_oos}. By varying the final machine learning classifier in our fused large language model using both fundamentals and textual self-descriptions, we yield an AUROC of 81.76\,\% (logistic regression), 82.51\,\% (elastic net), 81.75\,\% (random forest), and {82.78\,\% }(neural network). We observe a similar pattern with regard to the other performance metrics. For instance, the neural network achieves a {7.23-fold} ROI. Hence, the best overall AUROC is obtained by the neural network, followed by the elastic net, {logistic regression}, and the random forest. Altogether, this demonstrates the efficacy of our fused large language model based on a neural network in predicting startup success from VC platforms.

\begin{table}[H]
	\TABLE 
	{Out-of-sample performance of {different final machine learning classifiers within our fused large language model}. \label{tbl:prediction_oos}}
	{
	\OneAndAHalfSpacedXI
	\tiny
        \renewcommand{\arraystretch}{1.2}
	\sisetup{round-mode=places,round-precision=2,detect-weight,mode=text,table-column-width=1.2cm}
%	\sisetup{round-mode=places,round-precision=2}
	%\begin{tabular}{llllll}
        {
	\begin{tabular}{l>{\centering\arraybackslash}p{1.22cm}>{\centering\arraybackslash}p{1.15cm}>{\centering\arraybackslash}p{1.15cm}>{\centering\arraybackslash}p{1.15cm}>{\centering\arraybackslash}p{1.15cm}>{\centering\arraybackslash}p{1.15cm}>{\centering\arraybackslash}p{1.3cm}}
	  \toprule
	{\textbf{Classifier}} & {\textbf{Balanced accuracy}} & {\textbf{Precision}} & {\textbf{Recall}} & {\textbf{$\boldsymbol{F_1}$-score}} & {\textbf{AUROC}} &{\textbf{AUCPR}}  & {\textbf{ROI}} \\  
	\midrule 
  Logistic Regression & 73.71 (0.40) & 58.51 (0.47) & 78.79 (0.48) & 67.15 (0.43) & 81.76 (0.31) & 72.18 (0.59) &  704.98 (6.43) \\
  \addlinespace
  Elastic net & 74.27 (0.21) & 58.61 (0.31) & \textbf{80.43} (0.22) & \textbf{67.81} (0.22) & 82.51 (0.25) &  73.11 (0.50) &    706.30 (4.30) \\ 
  \addlinespace
  Random forest & 73.69 (0.64) &  58.28 (0.90) & 79.28 (1.68) &  67.16 (0.70) & 81.75 (0.63) & 72.17 (0.82) & 701.72 (12.37) \\
  \midrule
	Our fused large language model (neural network) & \textbf{74.33} (0.25) & \textbf{59.83} (1.79) & 78.28 (2.63) & 67.77 (0.15) & \textbf{82.78} (0.25) &  \textbf{73.70} (0.49) & \textbf{723.09} (24.56) \\
	 \bottomrule
	\end{tabular}
	}}
	{\hspace{-0.5cm} {\emph{Note:} Reported is the mean (and standard deviation) out-of-sample prediction performance across 5 random splits (in \%). The best value per metric and model is highlighted in bold.}
 \vspace{-.5cm}
 }
\end{table}

\subsection{{Sensitivity to Fine-Tuning Our Large Language Model}}\label{sec:finetuning}

{
We now experiment with fine-tuning our fused large language model. Specifically, we add a classification head to the $\text{[CLS]}$ embedding (i.e., $o_{\text{[CLS]}}$) for classifying startup success on top of BERT. We concatenate the fundamental variables to the $\text{[CLS]}$ embedding before feeding them to the classification head. This way, both the classification head is trained and BERT is fine-tuned simultaneously based on the task of predicting startup success. {Hence, the difference to no fine-tuning lies in the fact that we now allow for parameters in BERT to be fine-tuned for the task of classification.}

We fine-tuned BERT using the transformers framework from Huggingface \citep{Wolf.2020}. We use a training batch size of $32$ and a learning rate of $4 \cdot 10^{-5}$. We freeze the first 8 layers as they capture language patterns and an understanding of text in general. We update the weights of BERT and the classification head using the AdamW optimizer \citep{Loshchilov.2018}. We fine-tune for a maximum number of 5 epochs. We validate the performance every 50 steps. We performed early stopping when the loss on the hold-out set does not decrease for more than 5 steps.

\Cref{tbl:prediction_finetuning} reports the results. Overall, we do not observe any performance improvement when fine-tuning our fused-large language model. The performance of fine-tuning is comparable to that of our fused large language model with a neural net classifier. Specifically, fine-tuning our fused large language model yields a {0.12} percentage point decrease in accuracy, {0.1} percentage point decrease in AUROC, and {10.64} percentage point decrease in ROI, as compared to our not fine-tuned fused large language model. Hence, the pre-trained embeddings of BERT already capture textual information relevant to the task of success prediction. {Our findings underline an important aspect of machine learning: Increasing the number of trainable (or fine-tunable) parameters does not necessarily guarantee performance improvements. We discuss the finding later in \Cref{sec:discussion}.}

}

\begin{table}[H]
	\TABLE 
	{Prediction performance of our large language model with and without fine-tuning. \label{tbl:prediction_finetuning}}
	{
	\OneAndAHalfSpacedXI
	\tiny
        \renewcommand{\arraystretch}{1.2}
	\sisetup{round-mode=places,round-precision=2,detect-weight,mode=text,table-column-width=1.2cm}
%	\sisetup{round-mode=places,round-precision=2}
	%\begin{tabular}{llllll}
        {
	\begin{tabular}{p{1.1cm}>{\centering\arraybackslash}p{1.22cm}>{\centering\arraybackslash}p{1.15cm}>{\centering\arraybackslash}p{1.15cm}>{\centering\arraybackslash}p{1.15cm}>{\centering\arraybackslash}p{1.15cm}>{\centering\arraybackslash}p{1.15cm}>{\centering\arraybackslash}p{1.2cm}}
	  \toprule
	{\textbf{Fine-tuning}} & {\textbf{Balanced accuracy}} & {\textbf{Precision}} & {\textbf{Recall}} & {\textbf{$\boldsymbol{F_1}$-score}} & {\textbf{AUROC}} &{\textbf{AUCPR}}  & {\textbf{ROI}} \\  
	\midrule 
	No & \textbf{74.33} (0.25) & \textbf{59.83} (1.79) & 78.28 (2.63) & \textbf{67.77} (0.15) & \textbf{82.78} (0.25) &  \textbf{73.70} (0.49) & \textbf{723.09} (24.56) \\
 \addlinespace
    Yes & 74.21 (0.79) & 59.06 (2.67) & \textbf{79.77} (4.84) & 67.72 (0.77) & 82.68 (0.35) & 73.43 (0.48) & 712.45 (36.77) \\
	 \bottomrule
	\multicolumn{8}{l}{$^\dagger$Value not defined due to division by zero (\ie, there is no successful class)} \\
 \multicolumn{8}{l}{FV = fundamental variables, TSD = textual self-descriptions (via document embeddings)}
	\end{tabular}
	}}
	{\hspace{-0.5cm} {\emph{Note:} Reported is the mean (and standard deviation) out-of-sample prediction performance across 5 random splits (in \%). The best value per metric and model is highlighted in bold.}
 \vspace{-.5cm}
 }
\end{table}

\subsection{Prediction Performance Across Business Sectors}

% prediction performance across sectors

We now perform a sensitivity analysis in which we compare how the prediction performance from our fused large language model varies across business sectors (see \Cref{tbl:Sectors_performance}). {In general, startup activities and outcomes vary significantly across business sectors \citep{konon2018business}. For example, the sector of \textsc{Information Technology} typically features better data coverage and a higher number of startups, potentially leading to better predictability. Motivated by these differences, we perform a sensitivity analysis to provide insights into the extent to which textual self-descriptions contribute to performance gains across sectors.} Here, we focus our evaluations on the implementation based on a neural network, \ie, the best-performing classifier. We compare high-level business sectors for easier interpretability (this is different from the fine-grained but sparse industries that are reported on Crunchbase and that we use as predictors). Overall, we find that the prediction performance is fairly robust. The AUROC varies from {72.04}\,\% (\textsc{Energy}) to {85.39}\,\% (\textsc{Industrials}). This thus confirms that our fused large language model allows for accurate predictions across all business sectors. {Furthermore, including textual self-descriptions improves the prediction performance across most business sectors. The only exceptions are the four sectors with the smallest number of data points (\textsc{Energy}, \textsc{Materials}, \textsc{Real Estate}, and \textsc{Utilities}). For these sectors, including textual self-descriptions does not lead to a performance improvement as compared to using only the fundamental variables for prediction. Also, the standard deviation in the prediction performance is higher across these sectors. This implies that a sufficient number of training observations is necessary to make accurate predictions from textual self-descriptions.}\footnote{{We also tested whether the lower prediction performance in these business sectors could stem from more diverse textual self-descriptions as compared to other business sectors. However, the representations of the textual self-descriptions are (a)~not more/less discriminatory for successful vs. non-successful startups across business sectors, and (b)~not more/less diverse across business sectors, suggesting that more diversity in self-descriptions within specific business sectors is not a factor for lower prediction performance.}}

%Nevertheless, we see some variation of small size across business sectors. {Notably, for the sectors \textsc{Energy}, \textsc{Materials}, \textsc{Real estate}, and \textsc{Utilities}, including the textual self-descriptions does not lead to a performance improvement as compared to using only the fundamental variables for prediction. However, variation in performance is expected as these business sectors have the smallest number of observations available for training. This is also reflected by the larger standard deviation in the prediction performance over the five runs.

% More importantly, we find that the prediction performance is improved when including textual self-descriptions and that this improvement is consistently achieved across all business sectors. This thus provides further supporting evidence that self-descriptions are of operational value.  

\begin{table}[H]
\TABLE
{Prediction performance across business sectors.\label{tbl:Sectors_performance}}
{
\OneAndAHalfSpacedXI
%\footnotesize
\tiny
\renewcommand{\arraystretch}{1.2}
\sisetup{detect-weight,mode=text}
\sisetup{round-mode=places,round-precision=2,table-column-width=1.3cm}
%\begin{tabular}{llllll}
{
\begin{tabular}{p{1.527cm}rl>{\centering\arraybackslash}p{1.22cm}>{\centering\arraybackslash}p{1.15cm}>{\centering\arraybackslash}p{1.15cm}>{\centering\arraybackslash}p{1.15cm}>{\centering\arraybackslash}p{1.15cm}>{\centering\arraybackslash}p{1.15cm}>{\centering\arraybackslash}p{1.45cm}}
  \toprule
{\textbf{Business sector}} & {\textbf{\textit{N}}} & {\textbf{Predictors}} & {\thead{\tiny\textbf{Balanced} \\ \tiny\textbf{accuracy}}} & {\textbf{Precision}} & {\textbf{Recall}} & {\textbf{$\boldsymbol{F_1}$-score}} & {\textbf{AUROC}} & {\textbf{AUCPR}} & {\textbf{ROI}}\\  
  \midrule
                 \textsc{energy} &    92 &         FV &   \textbf{65.20} (3.50) &          54.74 (6.89) & \textbf{90.23} (6.55) & \textbf{67.77} (4.82) & \textbf{77.41} (2.51) & \textbf{73.75} (2.68) &           653.04 (94.75) \\
                &             &   FV + TSD &          64.38 (3.87) & \textbf{55.56} (7.49) &          79.81 (6.86) &          65.23 (6.03) &          72.04 (3.94) &          63.88 (6.16) & \textbf{664.36} (103.05) \\
                \addlinespace
             \textsc{materials} &    63 &         FV & \textbf{71.51} (2.44) & \textbf{57.23} (5.79) & \textbf{93.59} (6.45) & \textbf{70.83} (4.83) & \textbf{81.83} (3.23) & \textbf{77.03} (4.71) &  \textbf{687.33} (79.72) \\
             &             &   FV + TSD &          67.57 (4.18) &          55.35 (5.94) &          83.76 (7.49) &           66.5 (5.54) &          78.62 (5.75) &          71.73 (8.79) &            661.5 (81.68) \\
             \addlinespace
           \textsc{industrials} & 1412 &         FV &           73.00 (0.92) &          61.29 (2.18) &           84.57 (4.0) &          70.98 (0.74) &          82.85 (0.53) &          76.94 (1.41) &           743.17 (29.95) \\
            &        &   FV + TSD & \textbf{76.53} (0.77) & \textbf{65.56} (1.35) & \textbf{85.27} (1.84) & \textbf{74.11} (0.83) & \textbf{85.39} (0.41) & \textbf{79.98} (0.79) &   \textbf{801.92} (18.60) \\
            \addlinespace
\textsc{consumer discretionary} & 1382 &         FV &          72.19 (2.17) &          54.06 (3.64) & \textbf{77.33} (4.62) &          63.44 (1.55) &          80.26 (1.29) &          67.63 (1.66) &           643.66 (50.13) \\
&          &   FV + TSD & \textbf{74.32} (1.16) & \textbf{58.19} (2.32) &          75.51 (5.37) & \textbf{65.57} (1.15) & \textbf{82.52} (1.23) & \textbf{71.54} (1.15) &  \textbf{700.48} (31.85) \\
\addlinespace
      \textsc{consumer staples} &  642  &         FV &          68.47 (1.57) &           56.48 (2.4) & \textbf{83.55} (4.69) &          67.28 (0.85) &          78.09 (1.57) &           72.5 (1.93) &           677.02 (33.02) \\
       &           &   FV + TSD & \textbf{71.53} (2.25) & \textbf{60.52} (1.95) &          81.07 (4.84) & \textbf{69.25} (2.43) &   \textbf{80.70} (2.30) & \textbf{75.67} (1.63) &  \textbf{732.62} (26.88) \\
       \addlinespace
           \textsc{health care} &  658 &         FV &          63.69 (2.08) &          60.67 (2.93) & \textbf{92.39} (2.14) &          73.18 (1.63) &           77.09 (1.4) &           77.47 (2.2) &           734.59 (40.37) \\
           &            &   FV + TSD & \textbf{66.58} (1.19) & \textbf{63.47} (2.03) &          88.12 (1.46) & \textbf{73.77} (1.18) & \textbf{78.36} (0.46) & \textbf{78.88} (1.71) &   \textbf{773.20} (27.91) \\
           \addlinespace
            \textsc{financials} &  404 &         FV &          70.02 (2.24) &           58.92 (3.1) & \textbf{84.33} (3.73) &          69.25 (1.28) &          80.18 (1.18) &          76.38 (1.04) &           710.51 (42.59) \\
            &         &   FV + TSD & \textbf{72.73} (1.34) & \textbf{62.88} (2.56) &          81.94 (4.55) & \textbf{71.03} (0.98) & \textbf{82.84} (1.24) & \textbf{79.25} (1.68) &    \textbf{765.00} (35.20) \\
            \addlinespace
\textsc{information technology} & 2202 &         FV &          70.43 (1.43) &          58.07 (2.38) & \textbf{82.09} (4.08) &          67.92 (0.96) &           79.72 (0.9) &           73.5 (1.29) &           698.91 (32.81) \\
 &      &   FV + TSD &  \textbf{73.70} (0.73) & \textbf{62.44} (1.57) &          81.21 (2.56) &  \textbf{70.56} (0.90) &  \textbf{82.30} (0.51) & \textbf{76.64} (0.92) &  \textbf{759.02} (21.64) \\
 \addlinespace
\textsc{communication services} &  1747 &         FV &           72.36 (0.90) &          53.36 (4.16) & \textbf{77.42} (6.12) &          62.88 (1.37) &          79.96 (0.72) &          65.84 (1.07) &           634.13 (57.22) \\
 &            &   FV + TSD &  \textbf{73.40} (1.09) & \textbf{56.25} (2.55) &          74.52 (3.25) & \textbf{64.02} (0.83) & \textbf{81.46} (0.91) & \textbf{69.22} (1.13) &  \textbf{673.88} (35.07) \\
 \addlinespace
           \textsc{real estate} &  236 &         FV &          71.18 (3.56) &           53.46 (3.7) & \textbf{79.24} (3.45) & \textbf{63.74} (2.66) & \textbf{81.52} (2.89) &  \textbf{73.48} (3.80) &           635.41 (50.87) \\
            &           &   FV + TSD & \textbf{71.29} (2.03) &  \textbf{54.70} (3.42) &          75.98 (5.79) &          63.41 (1.97) &           80.70 (2.33) &          72.11 (2.32) &  \textbf{652.46} (47.02) \\
            \addlinespace
             \textsc{utilities} &    93 &         FV &          64.53 (5.13) &           59.6 (7.43) & \textbf{95.58} (2.67) &           73.12 (4.8) & \textbf{80.26} (1.93) & \textbf{79.26} (2.12) &          719.91 (102.22) \\
              &             &   FV + TSD & \textbf{68.64} (4.76) & \textbf{63.53} (6.34) &          90.71 (3.32) & \textbf{74.57} (4.67) &          78.55 (4.01) &          77.47 (5.68) &  \textbf{773.97} (87.28) \\
 \bottomrule
\multicolumn{7}{l}{FV = fundamental variables, TSD = textual self-descriptions} 
 
\end{tabular}
}}
{
\hspace{-0.5cm} {\emph{Note:} Reported is the mean (and standard deviation) out-of-sample prediction performance across 5 random splits (in \%). The best value per metric and sector is highlighted in bold.}
}
\end{table}

\subsection{Prediction Performance Across Investment Events}

We compare the prediction performance of our fused large language model across different events that are indicative of startup success, namely initial public offering, acquisition, and external funding. {For this, we evaluate our models on subsets of the out-of-sample test sets split by the different events. Hence, the corresponding accuracy quantifies, for example, to what extent startups are correctly classified in the subset of startups that eventually had an initial public offering.} We proceed analogously for acquisition and funding events. The results are reported in \Cref{tbl:class_performance_improvement}. Overall, the events vary in their frequency, as only a few startups had an initial public offering or had been acquired, whereas a larger proportion received external funding.

We find that the prediction performance is generally higher for initial public offerings and funding events. Here {82.05}\,\% of initial public offerings and {80.17}\,\% of funding events were predicted correctly. In contrast, only {65.54}\,\% of acquisitions were predicted correctly, implying that, for the latter, inferences are more challenging. Again, we confirm that machine learning benefits from incorporating textual self-descriptions. In fact, using textual self-descriptions increases the rate of correct classifications for initial public offerings by {4.02}\,\%, for acquisitions by {3.92}\,\%, and for funding events by {2.75}\,\%. Therefore, our findings suggest that textual self-descriptions from VC platforms are informative for predicting startup success, consistently across all success events.

\begin{table}[H]
\TABLE
{Prediction performance across different success events.  \label{tbl:class_performance_improvement}}
{
\OneAndAHalfSpacedXI
\tiny
\renewcommand{\arraystretch}{1.2}
{
\sisetup{detect-weight,mode=text}
%\sisetup{round-mode=places,round-precision=2,table-column-width=2.5cm,table-format=2.2}
%\begin{tabular}{llllll}
\begin{tabular}{lSl>{\centering\arraybackslash}p{1.2cm}} 
  \toprule
  \textbf{Success event}   & \textbf{\textit{N}} & \textbf{Predictors} & {\textbf{Balanced accuracy}}\\
	\midrule

Initial public offering & 17 & FV & 78.03 (5.33)   \\ 
   &  & FV + TSD & \textbf{82.05} (6.84)   \\ 
   \addlinespace 
Acquisition & 190 & FV & 61.62 (5.42)  \\ 
   &  & FV + TSD & \textbf{65.54} (1.36)  \\ 
   \addlinespace 
Funding & 1244 & FV & 77.42 (4.09) \\ 
   &  & FV + TSD & \textbf{80.17} (2.92)   \\ 
   \addlinespace 
Non-successful & 2584 & FV &  69.23 (6.27) \\ 
   &  & FV + TSD & \textbf{70.38} (3.10) \\ 
 \bottomrule
\multicolumn{4}{l}{FV = fundamental variables, TSD = textual self-descriptions}

\end{tabular}
}}
{
\hspace{-0.5cm} {\emph{Note:} Reported is the mean (and standard deviation) out-of-sample prediction performance across 5 random splits (in \%). Results are based on the neural network. The best value per success event is highlighted in bold. 
}}
\end{table}

\subsection{Robustness Checks}

We perform the following additional robustness checks. We evaluate the prediction performance of our fused large language model across different company characteristics (\ie, the age of a startup) and the length of the textual self-description. We find that the inclusion of textual self-descriptions improves the prediction performance considerably, which is consistent across startup ages and across different text lengths. This contributes to the robustness of our findings.

\subsubsection{Prediction Performance Across Startup Age.}
\label{apx:age_robustness}

The prediction performance with and without textual self-descriptions grouped across startup age is reported in \Cref{tbl:Age_performance}. For all age groups, the majority vote and random vote as na{\"i}ve baselines from machine learning are outperformed by a considerable margin and thus point toward the overall large prediction performance. In addition, all performance metrics increase by a considerable margin when including textual self-descriptions. This adds further robustness to our finding that textual self-descriptions are predictive of startup success. {Furthermore, the balanced accuracy is higher for older startups with and without textual self-descriptions, indicating that more established startups potentially yield more predictive information on Crunchbase.}

\begin{table}[H]
\TABLE
{Prediction performance across startup age. \label{tbl:Age_performance}}
{
\OneAndAHalfSpacedXI
\tiny
\renewcommand{\arraystretch}{1.2}
\sisetup{detect-weight,mode=text}
\sisetup{round-mode=places,round-precision=2,table-column-width=1.3cm}
%\begin{tabular}{llllll}
{
\begin{tabular}{lll>{\centering\arraybackslash}p{1.22cm}>{\centering\arraybackslash}p{1.15cm}>{\centering\arraybackslash}p{1.15cm}>{\centering\arraybackslash}p{1.15cm}>{\centering\arraybackslash}p{1.15cm}>{\centering\arraybackslash}p{1.15cm}>{\centering\arraybackslash}p{1.4cm}}
  \toprule
{\textbf{Startup age}} & {\textbf{\textit{N}}} & {\textbf{Predictors}} & {\thead{\tiny\textbf{Balanced} \\ \tiny\textbf{accuracy}}} & {\textbf{Precision}} & {\textbf{Recall}} & {\textbf{$\boldsymbol{F_1}$-score}} & {\textbf{AUROC}} & {\textbf{AUCPR}} & {\textbf{ROI}} \\  
  \midrule
 1--12 months & 1174 &         FV &          66.99 (1.56) &          58.82 (2.44) & \textbf{88.09} (2.86) &          70.47 (1.18) &          78.21 (1.25) &          75.34 (1.36) &          709.14 (33.51) \\
 & &   FV + TSD & \textbf{71.66} (0.73) & \textbf{64.55} (2.44) &          83.61 (4.01) & \textbf{72.76} (0.93) &  \textbf{81.20} (1.06) & \textbf{78.63} (0.61) & \textbf{788.05} (33.53) \\
 \addlinespace 
13--24 months &  1307 &         FV &          70.44 (1.93) &          54.77 (3.25) & \textbf{79.33} (4.78) &          64.68 (2.01) &          79.64 (0.75) &          70.76 (1.28) &          653.53 (44.75) \\
&  &   FV + TSD & \textbf{73.81} (0.85) & \textbf{59.62} (2.29) &           78.71 (3.2) & \textbf{67.78} (1.25) & \textbf{82.54} (1.09) &  \textbf{74.40} (1.38) &  \textbf{720.26} (31.50) \\
\addlinespace 
25--36 months & 1460 &         FV &          72.68 (1.53) &           53.9 (5.23) &          67.43 (8.08) &          59.41 (2.16) &          80.52 (0.82) &          63.87 (0.59) &           641.56 (71.9) \\
& &   FV + TSD & \textbf{74.05} (1.25) & \textbf{54.04} (1.99) & \textbf{70.24} (1.88) & \textbf{61.07} (1.67) & \textbf{82.15} (1.08) & \textbf{65.13} (1.89) &  \textbf{643.50} (27.44) \\
\bottomrule
\multicolumn{7}{l}{FV = fundamental variables, TSD = textual self-descriptions.} 
\end{tabular}
}}
{
\hspace{-0.5cm} {\emph{Note:} Reported is the mean (and standard deviation) out-of-sample prediction performance across 5 random splits (in \%). The best value per metric and age group is highlighted in bold.}
\vspace{-.5cm}
}
\end{table}

Varying the age of startups is also important for another reason: it allows us to assess the prediction performance across different time periods. Startups with an age between 1--12 months originate from 2015, an age between 13--24 months originate from 2014, etc. This thus contributes to the robustness of our findings.

\subsubsection{Prediction Performance Across Length of Textual Self-Description.}
\label{apx:text_robustness}

The prediction performance with and without textual self-descriptions across different lengths of the textual self-description is reported in \Cref{tbl:Length_performance}. For all length groups, a majority vote and a random vote are outperformed by machine learning. In addition, a clear improvement in AUROC is found for all groups when including textual self-descriptions. Overall, this adds robustness to our finding that textual self-descriptions are predictive of startup success. {Furthermore, the AUROC is higher for startups with longer textual self-descriptions. Similarly, both metrics increase for the baseline without textual self-descriptions. Still, the length of the textual self-description appears to play a minor role in the prediction performance.}

\begin{table}[H]
\TABLE
{Prediction performance for different lengths of the textual self-descriptions. \label{tbl:Length_performance}}
{
\OneAndAHalfSpacedXI
\tiny
\renewcommand{\arraystretch}{1.2}
\sisetup{detect-weight,mode=text}
\sisetup{round-mode=places,round-precision=2,table-column-width=1.3cm}
%\begin{tabular}{llllll}
{
\begin{tabular}{lll>{\centering\arraybackslash}p{1.22cm}>{\centering\arraybackslash}p{1.15cm}>{\centering\arraybackslash}p{1.15cm}>{\centering\arraybackslash}p{1.15cm}>{\centering\arraybackslash}p{1.15cm}>{\centering\arraybackslash}p{1.15cm}>{\centering\arraybackslash}p{1.4cm}}
  \toprule
{\textbf{Text length}} & {\textbf{\textit{N}}} & {\textbf{Predictors}} & {\thead{\tiny\textbf{Balanced} \\ \tiny\textbf{accuracy}}} & {\textbf{Precision}} & {\textbf{Recall}} & {\textbf{$\boldsymbol{F_1}$-score}} & {\textbf{AUROC}} & {\textbf{AUCPR}} & {\textbf{ROI}} \\  
  \midrule
50--100 words & 2306 &         FV &          71.05 (1.64) &          58.11 (3.79) &          79.47 (4.43) &          66.94 (1.11) &           79.82 (0.7) &           73.0 (0.72) &           699.47 (52.15) \\
 & &   FV + TSD & \textbf{73.55} (0.54) & \textbf{60.85} (1.55) & \textbf{80.59} (2.21) & \textbf{69.31} (0.63) & \textbf{82.29} (0.71) & \textbf{75.86} (1.25) &  \textbf{737.08} (21.28) \\
 \addlinespace 
101--200 words & 1103 &         FV &          73.56 (1.36) &          54.42 (2.63) &  \textbf{81.26} (4.10) &          65.06 (0.91) &          82.15 (0.69) &           70.65 (1.6) &           648.61 (36.25) \\
& &   FV + TSD & \textbf{75.23} (0.81) & \textbf{59.49} (2.85) &          76.32 (3.12) & \textbf{66.78} (1.46) & \textbf{83.36} (0.85) &  \textbf{71.90} (1.55) &  \textbf{718.42} (39.23) \\
\addlinespace 
   $\geq$201 words &  279 &         FV & \textbf{75.93} (2.65) &          38.74 (4.54) &  \textbf{76.80} (7.95) &           51.2 (3.88) &          83.36 (2.08) & \textbf{55.01} (9.07) &           432.95 (62.45) \\
   & &   FV + TSD &          73.46 (3.18) & \textbf{49.67} (7.84) &          59.47 (7.73) & \textbf{53.52} (4.87) & \textbf{84.04} (2.67) &          53.51 (5.41) & \textbf{583.33} (107.92) \\
\bottomrule
\multicolumn{7}{l}{FV = fundamental variables, TSD = textual self-descriptions}  
\end{tabular}
}}
{
\hspace{-0.5cm} {\emph{Note:} Reported is the mean (and standard deviation) out-of-sample prediction performance across 5 random splits (in \%). The best value per metric and length group is highlighted in bold.}
\vspace{-.5cm}
}
\end{table}

\subsection{Post-Hoc Explainability of Our Machine Learning Approach}
\label{sec:explanatory_analysis}

The previous analyses demonstrate the {performance improvement of including} textual self-descriptions for the task of startup success prediction. Now, we analyze the contributions of {each variable (\ie, fundamentals and textual self-descriptions)} for predicting startup success. To this end, we aim to understand how our fused large language model uses the variables to arrive at predictions. We use the SHAP value method \citep{Lundberg.2017}. Intuitively, the SHAP value method treats the prediction of a model as a cooperative game, i.e., the prediction (i.e., the payoff) must be allocated fairly among the feature values (i.e., the individual players) based on their contribution. Hence, the SHAP value method enables a nuanced understanding of how each feature contributes to the prediction of the model and is frequently used for understanding machine learning in management applications \citep[e.g.,][]{senoner2022using}.

SHAP values are computed for each observation separately, i.e., every feature within the vector of each observation is assigned a SHAP value. SHAP values can also be interpreted at the model level. Therefore, we quantify both feature attribution and feature importance based on the SHAP values. Feature attribution is directly determined by the SHAP values and feature importance is computed by averaging the absolute SHAP values across observations. {We follow previous research \citep{senoner2022using} and aggregate the SHAP values (sum) across the document embedding of the TSD to one feature representation}.

\begin{figure}[t!]
    \centering
    \begin{subfigure}{.715\textwidth}
        \centering
        \includegraphics[width=\linewidth]{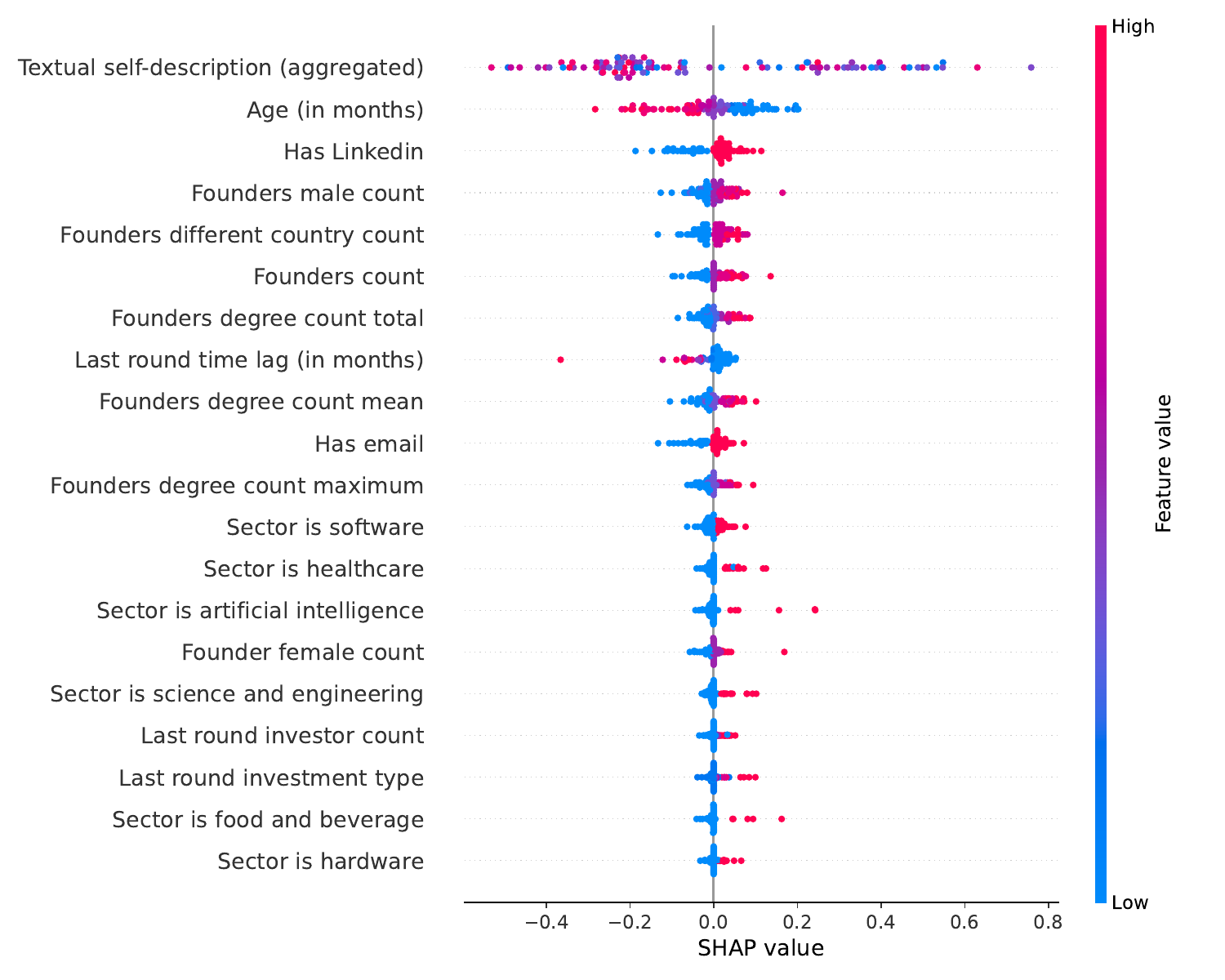} 
    \end{subfigure}%
    \begin{subfigure}{.285\textwidth}
        \centering
        \includegraphics[width=\linewidth]{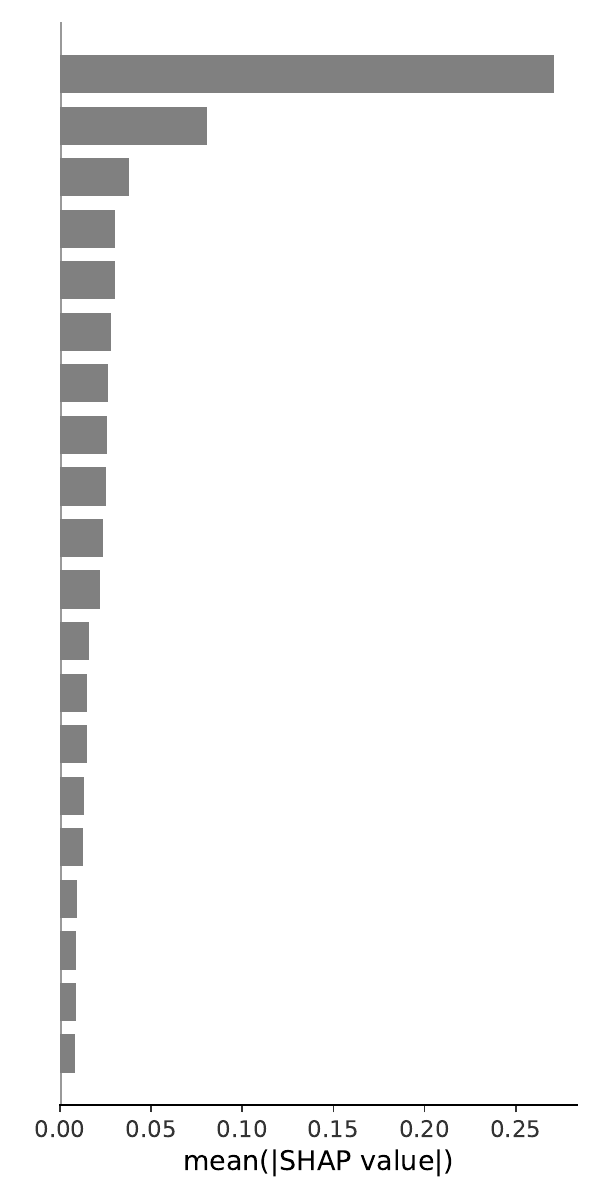} 
    \end{subfigure}%
    \caption{Left: Summary plot of the SHAP values for our fused large language model predictions. Each dot represents a SHAP value for a feature across different samples, where color indicates a high (red) to low (blue) feature value. Right: Bar plot of the mean (absolute) of SHAP values across all samples. The features are sorted by their feature importance.}
    \label{fig:SHAP}
    \vspace{-0.5cm}
\end{figure}

\Cref{fig:SHAP} shows the summary plot of SHAP values computed for the predictions of our fused large language model. {In the left plot}, the dots across each feature represent the feature attribution for each prediction of a specific observation. {The right plot shows the mean of the absolute SHAP values across all samples}. Both plots show the 20 features with the highest computed feature importance, ranking them from highest (top) to lowest (bottom) importance. Notably, {the aggregated representation of the textual self-description is the most important feature, indicating that it contributes, on average, the most to the prediction of our fused large language model. Here feature attributions range from $-0.53$ to $0.76$ with a mean absolute value of $0.29$. Thus, out of all features, the textual self-description adds the most to the predictions of our fused large language model}. 

{Variables characterizing the momentum of a startup also make important contributions to the predictions of our fused large language model. Overall, the age of the startup is the second most important feature. Feature attributions range from $-0.28$ for a startup age of 35.59 months to $0.20$ for a startup age of 4 months. On average, a higher age of startups is estimated to have a negative feature attribution to success prediction. This may indicate that startups with a longer market presence face reduced probabilities of success due to, for example, lower perceived growth potential and questionable viability of their business model if they have not yet achieved success. In addition, recent funding activities (last round time lag) are estimated as positive contributions to predictions of success, with feature attributions ranging from $-0.37$ to $0.05$. 
A plausible explanation might be that recent funding signals reduced risk, as other investors have recently found the startup promising enough to invest in (\ie, a form of validation). Similarly, the number of investors in the last investment round contributes positively to the success predictions, reinforcing the idea that previous funding may serve as a form of validation that predicts success also in the future.

Founder characteristics also play an important role in the predictions made by our fused large language model. Among these, the number of founders and their educational backgrounds are highly influential. Here, the total number of founders positively contributes to success predictions, which suggests that startups with a higher number of founders may benefit from diverse skill sets and shared responsibilities. The total number of degrees among the the founders also shows a positive contribution, with attributions ranging from $-0.09$ to $0.09$. This suggests that a higher number of educational degrees within the founding team may predict success, possibly reflecting the founding team's capability to tackle complex challenges and innovate. In addition, the presence of a LinkedIn profile (and an email) for founders also stands out as an important and positive contributor to the predictions of our model. This indicates that visible professional networking and the credibility it brings might be a strong predictor of later startup success. 

The contributions of the sectors in which a startup operates are also reflected in the SHAP values. For example, as seen by the SHAP values, \textsc{software} is the most influential sector feature. Startups in the \textsc{software} sector typically exhibit high growth potential, so that the information of whether a startup operates in this sector helps to predict later success. Similarly, sectors such as \textsc{healthcare} and \textsc{artificial intelligence} also show positive contributions, with \textsc{healthcare} ranging from $-0.04$ to $0.13$ and \textsc{artificial intelligence} from $-0.04$ to $0.22$. These sectors are typically characterized by innovative solutions of high impact. In contrast, sectors such as \textsc{food and beverage} and \textsc{hardware} show smaller but still positive contributions, which could be attributed to higher capital requirements and longer time to market.
}

%The inclusion of social media presence, specifically LinkedIn and email, also stands out as an important and positive contribution to the predictions of our model. Here, feature attributions for LinkedIn inclusion (and email inclusion) range from {−0.19} (and {−0.13}) to {0.11} (and {0.07}). . Moreover, a higher number of degrees held by the founders adds to the predicted probability of success, highlighting the contribution of the educational background to success probabilities. The tech sector, including high-growth areas like software and IT, along with broader new economy categories such as healthcare and financial services, shows a positive contribution. This is likely due to their potential for disruptive innovation, scalable business models, and favorable market trends.

\section{Discussion}
\label{sec:discussion}

\subsection{Managerial Implications}
% summary of findings

Our work demonstrates that VC platforms can be used to predict startup success and thus support investing decisions. We find that predictions from our fused large language model achieve an AUROC of up to {82.78}\,\%, a balanced accuracy of up to {74.33}\,\%, and a {7.33}-fold ROI. Thereby, baselines without machine learning (\eg, a majority vote) are outperformed by a considerable margin. Prior literature has already shown that various fundamental variables are predictors of startup success, whereas we show that additional predictive power is offered by textual self-descriptions. Here, we find that incorporating textual self-descriptions through our fused large language model increases the AUROC by {2.18} percentage points, the balanced accuracy by {2.33} percentage points, and the ROI by {52.25} percentage points. The increase in prediction performance is statistically significant. As such, our work is of direct managerial relevance as it provides computerized decision support for venture capitalists with the prospect of making financially rewarding investments. 

We also show that traditional machine learning methods for making predictions from text (e.g., bag-of-words {with manual feature extraction \citep{Kaiser.2020}}) are inferior to state-of-the-art methods based on large language models. {Traditional methods \citep[e.g.,][]{Kaiser.2020} rely on manually crafting features from text that might not capture the entire latent textual information. In contrast, our fused large language model utilizes so-called neural representation learning, capturing latent information in texts through an automated, data-driven procedure that learns from data.} {Notably, we observe that fine-tuning the language model does not increase the performance. The complexity behind the alignment of pre-trained knowledge and target domain characteristics has been discussed in recent NLP literature \citep{Bertsch.2024, Zhang.2024}, where evidence is provided that fine-tuning does not always help performance due to it being highly task- and data-specific. The fact that fine-tuning shows similar performance as no fine-tuning underlines an important aspect of machine learning: increasing the number of trainable or fine-tunable parameters does not necessarily guarantee performance improvements.} Thus managers should carefully consider the use of large language models when dealing with decision problems that involve text data. 

{The improvements in prediction performance when incorporating textual self-descriptions are robust across all business sectors and economically significant. To assess the practical implications, we translate the prediction performance into investment portfolio performance (ROI). Our results show significantly increased ROI when incorporating textual self-descriptions through our fused large language model: The best-performing baseline without textual self-descriptions amounts to a {6.71}-fold ROI, while our fused large language model achieves a {7.23}-fold ROI. The financial gains from our fused large language model can be further explained by the substantial costs of false positives in the context of startup investment decisions. False positive classifications for investment decisions lead to investing in startups predicted to succeed but ultimately failing. Hence, investments in startups that eventually fail lead to a potential loss of the entire investment amount. Our model significantly reduces the probability of false positives compared to the baselines, thereby increasing the overall returns from our machine learning approach for making investment decisions.}

{The Crunchbase database is widely used for academic research, which in turn yields practical implications. Crunchbase offers an online platform with comprehensive data on startups including fundamental variables (e.g., the age of the startup) and textual self-descriptions. Such data has key differences from the data traditionally collected by VC investors for decision-making \citep{kaplan2016venture,retterath2020benchmarking}. Here, two reasons stand out why investors traditionally have only little data about startup trajectories. On the one hand, investors typically collect only a few variables about startups (e.g., via scorecards) \citep{Bohm.2017,Yankov.2014} and often not in a structured format \citep{CFA.2015}. On the other hand, and more importantly, investors typically screen only a few dozen startups and thus only have access to startup data for a very small sample size \citep{retterath2020hit}, which precludes data-driven inferences. In sum, both of the aforementioned reasons are salient hurdles for training and deploying machine learning tools. As a remedy, prior literature evaluated the predictive ability of fundamental variables on Crunchbase \citep[e.g.,][]{Arroyo.2019, Dellermann.2017, Sharchilev.2018}. We add to prior literature by using large language models to incorporate the additional predictive ability of textual self-descriptions on Crunchbase. Hence, Crunchbase offers valuable data for VC investors and other practitioners regarding the evaluation of startups and the enhancement of decision-making tools.}

\subsection{Methodological Implications}

We contribute to business analytics research by demonstrating the operational value of large language models in the context of more effective investment decisions. Thereby, we connect to a growing stream of machine learning in business analytics \citep[e.g.,][]{Bastani.2021,Choi.2018,Cohen.2018,Misic.2020}. Different from explanatory analysis (\ie, regression analysis) that merely estimates associations in an in-sample setting, machine learning is concerned with how well inferences can be made in an out-of-sample setting. Here, we demonstrate an impactful application of machine learning in VC decision-making.   

% ein absatz zu den regression results

Large language models have several favorable advantages over traditional methods for natural language processing. On the one hand, large language models provide a flexible way to capture semantics and structure in textual materials, thereby bolstering the prediction performance over alternative machine learning approaches (e.g., bag-of-words). On the other hand, large language models can learn from vast amounts of unlabeled texts through pre-training. As such, large language models can often be applied out-of-the-box with little need for fine-tuning. This is beneficial as it greatly reduces the manual effort and the cost for data annotation. However, applications of large language models in business analytics are still rare, while we develop a tailored, \emph{fused} architecture for our decision-making problem. As shown above, large language models may need custom tailoring. In our case, we build a \emph{fused} large language model that can leverage running text but where the final prediction layer can also process structured data. As such, we expect that our \emph{fused} large language model is of direct relevance for many business analytics settings where the goal is to expand traditional operational information in structured form with additional text data.

% future research

Our study offers implications for the use of large language models in business analytics. We based our predictions on a tailored large language model, a recent innovation from machine learning research. We expect that large language models are beneficial for a wide array of managerial decision-making tasks. This opens new opportunities for research by adapting large language models to, for instance, sales and demand forecasting from social media data, credit scoring, and business failure prediction. 

\subsection{Limitations and Future Research}
\label{sec:limitations}

{As with other works, ours is not free of limitation, which offers promising directions for future research. First, large language models such as BERT may embed biases that are populated in downstream tasks. Large language models are trained on vast corpora of text data, which inevitably contain societal biases \citep{Bolukbasi.2016, Caliskan.2017, Garg.2018}. Consequently, there is a risk that these embedded biases could influence predictions \citep{de2022algorithmic}, potentially disadvantaging certain startups. Addressing this challenge requires ongoing efforts to mitigate biases within large language models. Future research could focus on refining these models to ensure equitable decision-making processes. For now, we call for careful use when deploying our model in practice. Second, our work is centered on data from the VC platform Crunchbase. While this choice is informed by prior research \citep{Arroyo.2019, Dellermann.2017, Sharchilev.2018}, it does introduce a limitation to our work. Crunchbase is a leading online VC platform that collects rich startup and investor data; however, it may not capture the full set of startups and investors globally. Future work might expand the data sources to include a broader spectrum of startups, enhancing the relevance and robustness across different sectors and regions. Third, the economic landscape of startups is dynamically evolving. To ensure ongoing predictive performance, continuous data collection and model retraining is needed. Lastly, the dynamic nature of the economic landscape might lead to startups adapting their textual self-descriptions in response to model predictions. This suggests an area for future research on the equilibrium implications of textual self-descriptions and model predictions.} {Specifically, analyzing equilibria could unveil the response of startups to prediction models in designing self-descriptions. Such analysis would require a different form of analysis using equilibria but not machine learning as in our paper.}

\section{Conclusion}
\label{sec:conclusion}

The majority of startups fail. Owing to this, the decision-making of investors is confronted with considerable challenges in identifying which startups will turn out to be successful. {To support investors in this task, we developed a tailored, \emph{fused} large language model that incorporates the textual self-description of startups alongside other fundamental variables to predict startup success. Here, we show that additional predictive power is offered by the textual self-descriptions.} Our model helps investors identify investment targets that promise financial returns. For this, our work provides computerized decision support that allows investors to automate their screening process with data-driven technologies. {Furthermore, our study highlights the potential of applying large language models in domains where relevant text data is available but has not traditionally been used for predicting outcomes needed for decision-making. For example, similar to ours, future work could attempt using textual self-descriptions of venture capitalists to predict the performance of their investments. In such scenarios, the findings from our study suggest that combining textual information with conventional data sources may have the potential to significantly enhance predictive accuracy and decision-making processes.}

\vspace{0.3cm}
{\footnotesize\textbf{Acknowledgments.} We thank Matthias Gey for help.}

  \newcommand{\dq}{"}
	\renewcommand{\textquotedbl}{"}
\OneAndAHalfSpacedXI
%\SingleSpacedXI
\bibliographystyle{informs2014} % outcomment this and next line in Case 1
{%\OneAndAHalfSpacedXI
\SingleSpacedXI

 \let\oldbibliography\thebibliography
 \renewcommand{\thebibliography}[1]{%
 	\oldbibliography{#1}%
 	\baselineskip7.5pt %Change this for line spacing within the same reference
 	\setlength{\itemsep}{2pt}% %Change this for spacing between two referneces
 }

\bibliography{literature} % if more than one, comma separated
}
\OneAndAHalfSpacedXI
\clearpage

\newpage

~~
\vspace{2cm}
\begin{center}
\Large\bfseries{Online Supplements}
\end{center}
\vspace{2cm}

\newpage

\begin{APPENDICES}

\section{Out-of-time Performance Evaluation}\label{appendix:outoftime_eval}

{We now repeat the analysis from \Cref{sec:results_basic} but now perform an out-of-time performance evaluation. Specifically, we split our data, ensuring that startups included in the test set originate from a period that follows the one represented by startups in the training set. Thereby, we can evaluate the ability of our model to predict the success of startups that stem from a period outside the one used for training. We proceed analogously as for the analysis in \Cref{sec:results_basic}, i.e., we perform a 10-fold cross validation and tune our hyperparameters via a randomized grid search (20 iterations) using the tuning grid from \Cref{tbl:tuning_grid}. \Cref{tbl:prediction_bow_oot} lists the results. Overall, our results remain robust, i.e., our fused large language model outperforms all baselines. Our fused large language model using both fundamentals and textual self-descriptions yields an AUROC of {78.91}\,\%, a balanced accuracy of {71.03}\,\%, and a {8.58}-fold ROI. However, this type of out-of-time splitting leads to smaller datasets, which increases the variance. Altogether, this demonstrates the efficacy of our fused large language model in predicting the success of startups that stem from a period outside the one represented in the training data.}

\begin{table}[H]
	\TABLE 
	{Out-of-time prediction performance of our large language model vs. the baselines. \label{tbl:prediction_bow_oot}}
	{
	\OneAndAHalfSpacedXI
	\footnotesize
        \renewcommand{\arraystretch}{1.2}
	\sisetup{round-mode=places,round-precision=2,detect-weight,mode=text,table-column-width=1.2cm}
%	\sisetup{round-mode=places,round-precision=2}
	%\begin{tabular}{llllll}
 {
	\begin{tabular}{ll>{\centering\arraybackslash}p{1.22cm}>{\centering\arraybackslash}p{1.15cm}>{\centering\arraybackslash}p{1.15cm}>{\centering\arraybackslash}p{1.15cm}>{\centering\arraybackslash}p{1.15cm}>{\centering\arraybackslash}p{1.15cm}>{\centering\arraybackslash}p{1.5cm}}
	  \toprule
	{\textbf{Approach}} & {\textbf{Predictors}} & {\textbf{Balanced accuracy}} & {\textbf{Precision}} & {\textbf{Recall}}  & {\textbf{$\boldsymbol{F_1}$-score}} & {\textbf{AUROC}} &{\textbf{AUCPR}} & {\textbf{ROI}}  \\ 
	\midrule 
  \multicolumn{8}{l}{Machine learning only with fundamental variables} \\
  \ldots logistic regression & FV & 65.46 & 61.93 &  83.70  & 71.19  & 74.41  & 74.42  & 752.04 \\
  \ldots elastic net & FV & 65.20  & 61.65  & 84.04  & 71.13  & 74.42  & 74.41  & 748.15 \\
  \ldots random forest & FV & 68.42  & 65.04  & 82.13  & 72.59  &  76.70  & 76.48  & 794.75  \\
  \ldots neural network & FV & 67.74 & 64.34 & 82.23 &  72.20 & 76.49 & 75.96 & 785.18  \\
  \midrule 
  \multicolumn{8}{l}{Manual feature extraction baseline \citep{Kaiser.2020}} \\ 
  \ldots logistic regression & FV + K\&K & 65.25  & 61.68  & 84.14  & 71.18  & 74.81  & 74.75  & 748.53  \\
  \ldots with elastic net & FV + K\&K & 65.17  & 61.55  & 84.53  & 71.23  & 74.82  & 74.76  & 746.71  \\
  \ldots with random forest 
  & FV + K\&K & 65.97  & 62.27  & 84.48  & 71.69  & 75.88  & 76.18  &  756.60  \\
	\ldots with neural network 
  & FV + K\&K & 67.23 & 63.37 & 84.68 & 72.49 & 76.84 &  76.3 & 771.79  \\
  \midrule 
  GloVe baseline \\   
  \ldots logistic regression & FV + GloVe & 66.65  & 63.15  & 82.97  & 71.72  & 76.03  & 76.13  & 768.79  \\
  \ldots with elastic net & FV + GloVe & 66.37  & 62.84  & 83.26  & 71.62  & 75.92  & 75.95  & 764.46  \\
  \ldots with random forest 
  & FV + GloVe & 66.58  & 62.56  & 85.95  & 72.41  & 77.48  & 77.29  & 760.62  \\
	\ldots with neural network  
  & FV + GloVe & 67.08 & 63.27 & 84.48 & 72.35 & 77.25 & 76.75 & 770.41  \\
  \midrule 
  Bag-of-words baseline \\ 
  \ldots logistic regression & FV + BOW & 65.96  & 62.76  &  81.6  & 70.95  & 74.85  & 75.09  & 763.45  \\
  \ldots with elastic net & FV + BOW & 65.63  & 62.45  & 81.64  & 70.77  & 74.88  &  75.10  & 759.12  \\
  \ldots with random forest  
  & FV + BOW & 67.01  & 63.36  &  83.70  & 72.12  & 77.35  & 77.42  & 771.61  \\
	\ldots with neural network 
  & FV + BOW & 66.08 & 62.18 & \textbf{85.61} & 72.03 & 76.79 & 76.56 & 755.36  \\
  \midrule 
  Our large language model	& FV + TSD & \textbf{71.03}  & \textbf{69.64}  & 76.11  & \textbf{72.73}  & \textbf{78.91}  & \textbf{78.59}  & \textbf{858.01} \\
	 \bottomrule
 \multicolumn{8}{l}{FV = fundamental variables, TSD = textual self-descriptions (via document embeddings)}
	\end{tabular}
	}}
	{\hspace{-0.5cm} {\emph{Note:} Reported is the mean (and standard deviation) out-of-sample prediction performance in \% across the 5 different splits. The best value per metric and model is highlighted in bold. K\&K is short for the features from \cite{Kaiser.2020}.}}

\end{table}

\section{Prediction performance of Final Machine Learning Classifier for the Baselines}
\label{appendix:long_eval}

{We now evaluate the performance of the final machine learning classifiers within our baselines in predicting startup success (see \Cref{tbl:prediction_bow_long}). Overall we observe some variation as to which final machine learning classifier performs best for each baseline. Specifically, using only the fundamental variables or incorporating the bag-of-words approach the random forest classifier performs best (AUROC: $80.95$ or $81.34$). For all other baselines, the neural network consistently outperforms the other final machine learning classifiers. We attribute the variation in the best-performing final machine learning classifier to the fact that the input size and the complexity vary for each baseline.}

\begin{table}[H]
	\TABLE 
	{Prediction performance of final machine learning classifier for the baselines. \label{tbl:prediction_bow_long}}
	{
	\OneAndAHalfSpacedXI
	\tiny
        \renewcommand{\arraystretch}{1.2}
	\sisetup{round-mode=places,round-precision=2,detect-weight,mode=text,table-column-width=1.2cm}
 {
	\begin{tabular}{ll>{\centering\arraybackslash}p{1.22cm}>{\centering\arraybackslash}p{1.15cm}>{\centering\arraybackslash}p{1.15cm}>{\centering\arraybackslash}p{1.15cm}>{\centering\arraybackslash}p{1.15cm}>{\centering\arraybackslash}p{1.15cm}>{\centering\arraybackslash}p{1.2cm}}
	  \toprule
	{\textbf{Approach}} & {\textbf{Predictors}} & {\textbf{Balanced accuracy}} & {\textbf{Precision}} & {\textbf{Recall}}  & {\textbf{$\boldsymbol{F_1}$-score}} & {\textbf{AUROC}} &{\textbf{AUCPR}} & {\textbf{ROI}}  \\ 
	  \midrule
	Majority vote & --- & 50.00 & {\textemdash{$^\dagger$}} & 0.00 & {\textemdash{$^\dagger$}} & 50.00 & 0.00 & {\textemdash{$^\dagger$}} \\  
 Random vote & --- & 50.00 & 36.59 & 36.14 & 36.36 & 50.00 & 23.16 & 403.40 \\  
	\midrule 
  \multicolumn{8}{l}{Machine learning with fundamental variables only} \\
  \ldots logistic regression & FV & 71.20 (0.54) & 55.15 (0.44) & 78.06 (1.21) & 64.63 (0.61) & 79.47 (0.58) &  69.30 (0.57) &  658.67 (6.05) \\
  \ldots elastic net & FV & 71.24 (0.53) &  55.10 (0.46) &  78.30 (1.14) & 64.68 (0.58) &  79.50 (0.55) &  69.31 (0.60) &  658.06 (6.34) \\ 
  \ldots random forest & FV & 72.43 (0.47) & 56.12 (0.49) & 79.97 (1.51) & 65.95 (0.53) &  80.95 (0.60) & 71.33 (0.82) &   672.10 (6.73) \\
  \ldots neural network & FV & 72.00 (1.33) & 56.03 (3.27) & 79.46 (4.45) & 65.56 (0.92) &  80.60 (0.44) & 70.92 (0.68) & 670.84 (45.05) \\
  \midrule 
  \multicolumn{8}{l}{Manual feature extraction baseline \citep{Kaiser.2020}} \\
  \ldots logistic regression & FV + K\&K & 72.10 (0.42) &  55.80 (0.36) &  79.61 (1.10) & 65.61 (0.47) & 80.31 (0.61) & 69.94 (0.62) &  667.69 (5.01) \\
  \ldots with elastic net & FV + K\&K & 72.21 (0.34) & 55.87 (0.26) & 79.83 (1.02) & 65.73 (0.39) & 80.34 (0.58) & 69.94 (0.66) &  668.62 (3.62) \\ 
  \ldots with random forest  & FV + K\&K & 72.41 (0.55) & 55.82 (0.53) & 80.68 (1.16) & 65.98 (0.59) &  80.90 (0.49) & 71.34 (0.55) &  667.92 (7.22) \\	
  \ldots with neural network 
  & FV + K\&K & 72.40 (1.15) & 56.83 (3.13) & 78.86 (4.65) & 65.89 (0.77) & 81.22 (0.39) & 71.42 (0.63) & 681.88 (43.12) \\
  \midrule 
  GloVe baseline \\  
  \ldots logistic regression & FV + GloVe & 72.63 (0.65) & 56.62 (0.59) & 79.45 (1.04) &  66.12 (0.70) &  81.01 (0.60) & 71.08 (0.81) & 678.88 (8.09) \\
  \ldots with elastic net & FV + GloVe & 72.83 (0.50) & 56.75 (0.46) &  79.81 (0.80) & 66.33 (0.54) & 81.04 (0.57) & 70.97 (0.85) &  680.7 (6.34) \\

  \ldots with random forest 
  & FV + GloVe & 72.30 (0.41) & 55.49 (0.55) & 81.17 (0.91) & 65.92 (0.42) & 81.27 (0.64) & 71.58 (0.85) & 663.44 (7.52) \\

	\ldots with neural network 
  & FV + GloVe & 71.87 (1.93) & 53.86 (3.27) & \textbf{85.16} (3.72) & 65.85 (1.48) & 81.89 (0.59) & 72.59 (0.59) & 640.9 (44.93) \\
  \midrule 
  Bag-of-words baseline \\  
  \ldots logistic regression & FV + BOW & 72.17 (0.55) & 56.29 (0.68) & 78.64 (1.02) & 65.61 (0.57) & 80.32 (0.46) & 70.53 (0.79) &  674.33 (9.32) \\
  \ldots with elastic net & FV + BOW & 72.15 (0.48) & 56.19 (0.64) &  78.81 (0.90) &   65.60 (0.50) & 80.35 (0.47) & 70.53 (0.88) &  673.05 (8.78) \\
  \ldots with random forest  
  & FV + BOW & 72.82 (0.69) & 56.29 (0.87) &  80.98 (1.60) &  66.40 (0.71) & 81.34 (0.57) &  71.83 (0.60) & 674.36 (11.92) \\

	\ldots with neural network 
  & FV + BOW & 72.38 (0.49) & 55.71 (1.23) & 80.95 (1.49) & 65.98 (0.38) &  81.10 (0.18) & 71.74 (0.63) & 666.38 (16.88) \\
  \midrule 
  Our large language model	& FV + TSD & \textbf{74.33} (0.25) & \textbf{59.83} (1.79) & 78.28 (2.63) & \textbf{67.77} (0.15) & \textbf{82.78} (0.25) &  \textbf{73.70} (0.49) & \textbf{723.09} (24.56) \\
	 \bottomrule
  \multicolumn{7}{l}{$^\dagger$Value not defined due to division by zero (\ie, there is no successful class)} \\
 \multicolumn{5}{l}{FV = fundamental variables,}\\
   \multicolumn{5}{l}{TSD = textual self-description (via large language model),}\\
   \multicolumn{9}{l}{K\&K = manual feature extraction from \citet{Kaiser.2020} (from textual self-description),}\\
   \multicolumn{5}{l}{BOW = bag-of-words (from textual self-descriptions)}
	\end{tabular}
	}}
	{\hspace{-0.5cm} {\emph{Note:} Reported is the mean (and standard deviation) out-of-sample prediction performance across 5 random splits (in \%). The best value per metric and model is highlighted in bold. K\&K is short for the features from \citet{Kaiser.2020}.}}
 
\end{table}

\newpage
\section{Performance of Varying the Input Variables within our Fused Large Language Model}
\label{apx:input_vars}

{\Cref{tbl:prediction_oos_long} compares the performance of the final machine learning machine learning classifiers within our fused large language model for varying input variables. Specifically, we assess the relative gain from using textual self-descriptions. For this purpose, we compare the prediction performance with two specific sets of predictors: (a)~our machine learning approach trained only on fundamental variables and (b)~our machine learning approach trained on both fundamental variables and textual self-descriptions ($=$ our fused large language model). Across all machine learning classifiers, we find consistent evidence that the prediction performance is improved when considering textual self-descriptions. By including textual self-descriptions, the AUROC improves by {2.29 percentage points (logistic regression), 3.01 percentage points (elastic net), 1.8 percentage points (random forest), and 2.18 percentage points (neural network)}. The improvements in the balanced accuracy amount to {2.51 percentage points (logistic regression), 3.03 percentage points (elastic net), 1.26 percentage points (random forest), and 2.33 percentage points (neural network)}. 

{The increases in ROI amount to {46.31 percentage points (logistic regression), 48.24 percentage points (elastic net), 29.62 percentage points (random forest), and 52.25 percentage points (neural network). The increases in ROI highlight the economic value of incorporating textual self-descriptions when predicting startup success}.}

We also assess whether the improvement in prediction performance due to including textual self-descriptions is statistically significant. For this purpose, we utilize McNemar's test comparing the predictions with and without textual self-descriptions (while including fundamental variables). Here, we find that the performance increase is statistically significant at common significance thresholds for all considered machine learning classifiers, \ie, {logistic regression ($\chi^2$-statistic $= 21.70$; $p < 0.01$), elastic net ($\chi^2$-statistic $= 25.55$; $p < 0.01$), random forest {($\chi^2$-statistic $= 5.95$; $p < 0.05$)}, and neural network ($\chi^2$-statistic $= 11.00$; $p < 0.01$)}. In sum, the improvements from using textual self-descriptions are achieved consistently across all classifiers. The results thus confirm that textual self-descriptions have predictive power and thus are of operational value.

% text only
{
For comparison, we also report the prediction performance of machine learning that is fed solely with textual self-descriptions. Here, the majority vote and random vote as na{\"i}ve baselines are again outperformed by a considerable margin. Furthermore, we observe that the prediction performance of using only textual self-descriptions is comparable but slightly inferior to the prediction performance obtained by using only fundamental variables. For instance, for the neural network, the AUROC is {80.60}\,\% for a model with only fundamental variables vs. an AUROC of {77.24}\,\% for a model with only textual self-descriptions. 
}
}

\begin{table}[H]
	\TABLE 
	{{Out-of-sample performance for varying the input variables within our fused large language model}. \label{tbl:prediction_oos_long}}
	{
	\OneAndAHalfSpacedXI
	\tiny
        \renewcommand{\arraystretch}{1.2}
	\sisetup{round-mode=places,round-precision=2,detect-weight,mode=text,table-column-width=1.2cm}
%	\sisetup{round-mode=places,round-precision=2}
	%\begin{tabular}{llllll}
        {
	\begin{tabular}{ll>{\centering\arraybackslash}p{1.22cm}>{\centering\arraybackslash}p{1.15cm}>{\centering\arraybackslash}p{1.15cm}>{\centering\arraybackslash}p{1.15cm}>{\centering\arraybackslash}p{1.15cm}>{\centering\arraybackslash}p{1.15cm}>{\centering\arraybackslash}p{1.3cm}}
	  \toprule
	{\textbf{Classifier}} & {\textbf{Predictors}} & {\textbf{Balanced accuracy}} & {\textbf{Precision}} & {\textbf{Recall}} & {\textbf{$\boldsymbol{F_1}$-score}} & {\textbf{AUROC}} &{\textbf{AUCPR}}  & {\textbf{ROI}} \\  
	\midrule 
  Logistic Regression & FV & 71.20 (0.54) & 55.15 (0.44) & 78.06 (1.21) & 64.63 (0.61) & 79.47 (0.58) &  69.3 (0.57) &  658.67 (6.05) \\
  & TSD & 68.91 (0.28) & 52.99 (0.43) & 75.37 (0.45) & 62.23 (0.24) & 76.06 (0.46) & 63.85 (0.98) & 628.97 (5.95) \\ 
   & FV + TSD & \textbf{73.71} (0.40) & \textbf{58.51} (0.47) & \textbf{78.79} (0.48) & \textbf{67.15} (0.43) & \textbf{81.76} (0.31) & \textbf{72.18} (0.59) &  \textbf{704.98} (6.43) \\
	\midrule 
  Elastic net & FV & 71.24 (0.53) &  55.1 (0.46) &  78.30 (1.14) & 64.68 (0.58) &  79.5 (0.55) &  69.31 (0.60) &  658.06 (6.34) \\
  & TSD & 69.14 (0.20) &  53.02 (0.40) & 76.18 (0.74) & 62.52 (0.18) & 76.46 (0.35) &  64.42 (0.70) & 629.43 (5.54) \\  
   & FV + TSD & \textbf{74.27} (0.21) & \textbf{58.61} (0.31) & \textbf{80.43} (0.22) & \textbf{67.81} (0.22) & \textbf{82.51} (0.25) &  \textbf{73.11} (0.50) &    \textbf{706.30} (4.30) \\ 
	\midrule 
  Random forest & FV & 72.43 (0.47) & 56.12 (0.49) & 79.97 (1.51) & 65.95 (0.53) &  80.95 (0.6) & 71.33 (0.82) &   672.10 (6.73) \\
  & TSD & 66.40 (0.33) & 48.48 (0.16) & \textbf{81.34} (1.01) & 60.75 (0.38) & 74.62 (0.37) & 62.74 (0.19) & 566.94 (2.18) \\ 
   & FV + TSD & \textbf{73.69} (0.64) &  \textbf{58.28} (0.90) & 79.28 (1.68) &  \textbf{67.16} (0.70) & \textbf{81.75} (0.63) & \textbf{72.17} (0.82) & \textbf{701.72} (12.37) \\
		\midrule
	Neural network & FV & 72.00 (1.33) & 56.03 (3.27) & \textbf{79.46} (4.45) & 65.56 (0.92) &  80.60 (0.44) & 70.92 (0.68) & 670.84 (45.05) \\
  & TSD & 69.08 (0.58) & 52.24 (1.57) & 78.69 (3.45) & 62.73 (0.41) &  77.24 (0.4) &  65.5 (0.64) & 618.73 (21.59) \\ 
   & FV + TSD & \textbf{74.33} (0.25) & \textbf{59.83} (1.79) & 78.28 (2.63) & \textbf{67.77} (0.15) & \textbf{82.78} (0.25) &  \textbf{73.70} (0.49) & \textbf{723.09} (24.56) \\
	 \bottomrule
	\multicolumn{9}{l}{$^\dagger$Value not defined due to division by zero (\ie, there is no successful class)} \\
 \multicolumn{9}{l}{FV = fundamental variables, TSD = textual self-descriptions (via document embeddings)}
	\end{tabular}
	}}
	{\hspace{-0.5cm} {\emph{Note:} Reported is the mean (and standard deviation) out-of-sample prediction performance across 5 random splits (in \%). The best value per metric and model is highlighted in bold.}}
\end{table} 

\section{Cross-Correlation of Fundamental Variables}
\label{apx:corr_fund}

The cross-correlation of the fundamental variables is shown in \Cref{tbl:cors_fundamental}. Strong correlations are observed, for example, between variables indicating the number of degrees of the founders (\eg, founders degree count maximum and mean). While collinearity might affect correct estimates in explanatory analysis, it is beneficial for machine learning. The reason is that strong correlations often yield more powerful classifiers \citep{Hastie.2009}.

\clearpage \newpage
\begin{sidewaystable}[ph!]
\TABLE
{Cross-correlations of fundamental variables.\label{tbl:cors_fundamental}}
{
\OneAndAHalfSpacedXI
\scriptsize
\sisetup{round-mode=places,round-precision=2}
\begin{tabular}{rl *{21}{S[table-format=2.2,table-column-width=.5cm]}S[table-column-width=.5cm, table-format=1.0]}
\toprule
& \textbf{Variable} & \mc{$1$} & \mc{$2$}  & \mc{$3$} & \mc{$4$} & \mc{$5$} & \mc{$6$} & \mc{$7$} & \mc{$8$} & \mc{$9$} & \mc{$10$} & \mc{$11$} & \mc{$12$} & \mc{$13$} & \mc{$14$} & \mc{$15$} & \mc{$16$} & \mc{$17$} & \mc{$18$} & \mc{$19$} & \mc{$20$} & \mc{$21$} & \mc{$22$}\\
\midrule 
  1 & Age & {1} &  &  &  &  &  &  &  &  &  &  &  &  &  &  &  &  &  &  &  &  &  \\ 
    2 & Has email & -0.02 & {1} &  &  &  &  &  &  &  &  &  &  &  &  &  &  &  &  &  &  &  &  \\ 
    3 & Has phone &  0.03 &  0.29 & {1} &  &  &  &  &  &  &  &  &  &  &  &  &  &  &  &  &  &  &  \\ 
    4 & Has Facebook &  0.05 &  0.22 &  0.11 & {1} &  &  &  &  &  &  &  &  &  &  &  &  &  &  &  &  &  &  \\ 
    5 & Has Twitter &  0.07 &  0.21 &  0.06 &  0.54 & {1} &  &  &  &  &  &  &  &  &  &  &  &  &  &  &  &  &  \\ 
    6 & Has LinkedIn & -0.10 &  0.19 &  0.12 &  0.23 &  0.28 & {1} &  &  &  &  &  &  &  &  &  &  &  &  &  &  &  &  \\ 
    7 & Number of investment rounds &  0.24 &  0.01 & -0.04 &  0.05 &  0.05 &  0.02 & {1} &  &  &  &  &  &  &  &  &  &  &  &  &  &  &  \\ 
    8 & Raised funding &  0.03 &  0.01 &  0.04 & -0.01 & -0.01 &  0.02 &  0.14 & {1} &  &  &  &  &  &  &  &  &  &  &  &  &  &  \\ 
    9 & Last round raised funding &  0.01 &  0.00 &  0.03 & -0.03 & -0.02 &  0.01 &  0.07 &  0.93 & {1} &  &  &  &  &  &  &  &  &  &  &  &  &  \\ 
   10 & Last round post money evaluation &  0.01 &  0.01 &  0.01 &  0.01 &  0.01 &  0.01 &  0.07 &  0.10 &  0.02 & {1} &  &  &  &  &  &  &  &  &  &  &  &  \\ 
   11 & Last round time lag &  0.32 & -0.08 & -0.06 & -0.02 & -0.04 & -0.11 &  0.68 &  0.06 &  0.05 &  0.02 & {1} &  &  &  &  &  &  &  &  &  &  &  \\ 
   12 & Investor count &  0.16 &  0.06 & -0.02 &  0.08 &  0.09 &  0.09 &  0.66 &  0.17 &  0.06 &  0.09 &  0.33 & {1} &  &  &  &  &  &  &  &  &  &  \\ 
   13 & Last round investor count &  0.10 &  0.03 & -0.04 &  0.04 &  0.06 &  0.05 &  0.48 &  0.09 &  0.10 &  0.02 &  0.38 &  0.62 & {1} &  &  &  &  &  &  &  &  &  \\ 
   14 & Known investor count &  0.22 &  0.00 & -0.04 &  0.04 &  0.04 &  0.01 &  0.91 &  0.15 &  0.08 &  0.06 &  0.70 &  0.69 &  0.55 & {1} &  &  &  &  &  &  &  &  \\ 
   15 & Last round known investor count &  0.22 & -0.01 & -0.05 &  0.04 &  0.03 &  0.00 &  0.92 &  0.14 &  0.09 &  0.05 &  0.74 &  0.63 &  0.57 &  0.98 & {1} &  &  &  &  &  &  &  \\ 
   16 & Founders count &  0.02 &  0.11 & -0.09 &  0.12 &  0.19 &  0.17 &  0.26 &  0.05 &  0.03 &  0.02 &  0.11 &  0.26 &  0.20 &  0.25 &  0.24 & {1} &  &  &  &  &  &  \\ 
   17 & Founders different country count &  0.01 &  0.10 & -0.09 &  0.12 &  0.18 &  0.16 &  0.23 &  0.03 &  0.02 &  0.01 &  0.09 &  0.22 &  0.17 &  0.21 &  0.21 &  0.91 & {1} &  &  &  &  &  \\ 
   18 & Founders male count &  0.02 &  0.10 & -0.08 &  0.11 &  0.18 &  0.17 &  0.26 &  0.06 &  0.04 &  0.02 &  0.11 &  0.26 &  0.20 &  0.25 &  0.24 &  0.97 &  0.88 & {1} &  &  &  &  \\ 
   19 & Founders female count &  0.00 &  0.09 & -0.08 &  0.12 &  0.17 &  0.13 &  0.18 &  0.02 &  0.01 &  0.01 &  0.06 &  0.18 &  0.15 &  0.17 &  0.17 &  0.79 &  0.82 &  0.64 & {1} &  &  &  \\ 
   20 & Founders degree count total &  0.01 &  0.10 & -0.07 &  0.11 &  0.16 &  0.18 &  0.25 &  0.06 &  0.03 &  0.02 &  0.10 &  0.27 &  0.21 &  0.24 &  0.24 &  0.77 &  0.71 &  0.74 &  0.63 & {1} &  &  \\ 
   21 & Founders degree count maximum &  0.01 &  0.10 & -0.07 &  0.11 &  0.17 &  0.18 &  0.24 &  0.05 &  0.02 &  0.02 &  0.09 &  0.25 &  0.19 &  0.23 &  0.22 &  0.78 &  0.79 &  0.75 &  0.70 &  0.94 & {1} &  \\ 
   22 & Founders degree count mean &  0.00 &  0.10 & -0.07 &  0.11 &  0.17 &  0.18 &  0.23 &  0.04 &  0.02 &  0.02 &  0.09 &  0.24 &  0.19 &  0.22 &  0.22 &  0.78 &  0.80 &  0.75 &  0.71 &  0.91 &  0.99 & {1} \\  

\bottomrule
\multicolumn{20}{l}{Stated: Pearson correlation coefficient}
& \multicolumn{4}{r}{$N$ = 20,172}
\end{tabular}
}
{\SingleSpacedXI\footnotesize

}
\end{sidewaystable}
\clearpage

\end{APPENDICES}

\end{document}